\documentclass[journal]{IEEEtran}
\usepackage{graphicx}
\usepackage{multirow}
\usepackage{epstopdf}
\usepackage{times}
\usepackage{amsmath}
\usepackage{amssymb}
\usepackage{amsfonts}
\usepackage{color}
\usepackage{subfigure}
\usepackage[colorlinks]{hyperref}
 \usepackage{colortbl}
\usepackage[ruled, linesnumbered, vlined]{algorithm2e}

\usepackage{algorithmic}
\usepackage{rotating}
\usepackage{adjustbox,lipsum}
\usepackage{dsfont}
\usepackage{booktabs}
\usepackage{multirow}
\usepackage{array}

\usepackage{pifont}

\newcommand{\etal}{{\em et al.\,}}       % et al.
\newcommand{\ie}{{\em i.e.}}           % i.e.

\newcommand{\bl}[1]{\textcolor{blue}{#1}}

\begin{document}

\title{Generalizable Metric Network for Cross-domain Person Re-identification}

\author{Lei Qi,
        Ziang Liu,
        Yinghuan Shi,
        Xin Geng
%\thanks{Manuscript received August 13, 2019; revised November 01, 2019;
%accepted March 21, 2020. Date of publication  XXXX, 2020; date of
%current version March 24, 2020. This work was supported by NSFC (61806092, 61432008, 61673203) and Jiangsu Natural Science Foundation (BK20180326).  This paper was recommended by Associate Editor Dr. Yap-Peng Tan. (Corresponding author: Yang Gao.)}
%\thanks{This work was supported by NSFC Program (62206052, 62125602, 62076063, 62222604), CAAI-Huawei MindSpore Project (CAAIXSJLJJ-2021-042A), China Postdoctoral Science Foundation Project (2021M690609), Jiangsu Natural Science Foundation Project (BK20210224), and the Big Data Computing Center of Southeast University.}
\thanks{The work is supported by NSFC Program (Grants No. 62206052, 62125602, 62076063), Jiangsu Natural Science Foundation Project (Grant No. BK20210224), and the Xplorer Prize.}
\thanks{Lei Qi, Ziang Liu and Xin Geng are with the School of Computer Science and Engineering, Southeast University, and Key Laboratory of New Generation Artificial Intelligence Technology and Its Interdisciplinary Applications (Southeast University), Ministry of Education, China, 211189 (e-mail: qilei@seu.edu.cn; liuziang@seu.edu.cn; xgeng@seu.edu.cn).}
\thanks{Yinghuan Shi is with the State Key Laboratory for Novel Software Technology, Nanjing University, Nanjing, China, 210023 (e-mail: syh@nju.edu.cn).}
\thanks{Corresponding author: Xin Geng.}
% <-this % stops a space
%\thanks{Manuscript received O 10, 2017; revised August 26, 2017.}
}


\markboth{ }%
{Shell \MakeLowercase{\textit{et al.}}: Bare Demo of IEEEtran.cls for IEEE Journals}

\maketitle

\begin{abstract}
Person Re-identification (Re-ID) is a crucial technique for public security and has made significant progress in supervised settings. However, the cross-domain (\ie, domain generalization) scene presents a challenge in Re-ID tasks due to unseen test domains and domain-shift between the training and test sets. To tackle this challenge, most existing methods aim to learn domain-invariant or robust features for all domains. In this paper, we observe that the data-distribution gap between the training and test sets is smaller in the sample-pair space than in the sample-instance space. Based on this observation, we propose a Generalizable Metric Network (GMN) to further explore sample similarity in the sample-pair space. Specifically, we add a Metric Network (M-Net) after the main network and train it on positive and negative sample-pair features, which is then employed during the test stage. Additionally, we introduce the Dropout-based Perturbation (DP) module to enhance the generalization capability of the metric network by enriching the sample-pair diversity. Moreover, we develop a Pair-Identity Center (PIC) loss to enhance the model's discrimination by ensuring that sample-pair features with the same pair-identity are consistent. We validate the effectiveness of our proposed method through a lot of experiments on multiple benchmark datasets and confirm the value of each module in our GMN.
\end{abstract}


% \fi
\IEEEpeerreviewmaketitle

%\vspace*{-10pt}% µ÷Õû¼ä¾à
%%%%%%%%% BODY TEXT
\section{Introduction}
\IEEEPARstart{A}{s} the demand for video surveillance applications grows, person re-identification (Re-ID) has become an essential technique in public security to identify individuals from videos captured by different cameras~\cite{qi2021adversarial,qi2022novel}. Person Re-ID has gained attention from academia and industry due to its potential in various applications~\cite{zheng2016person,ye2020deep,DBLP:journals/tcsv/LengYT20,qi2020progressive}, where images of the same person are matched across non-overlapping camera views. However, Re-ID is a difficult task due to variations in body pose, occlusion, illumination, image resolution, background, and other factors~\cite{DBLP:journals/pami/LiZG20,chen2021occlude,DBLP:journals/tcsv/QiWHSG20,DBLP:journals/tifs/MaJZTP20}. Person Re-ID can be considered as a special case of the image retrieval task, where the goal is to find these images from a large gallery set, which have the same identity with a given query image.

% \IEEEPARstart{P}{e}rson re-identification (Re-ID) is a crucial technique in public security as it can efficiently aid in identifying specific individuals from large-scale videos~\cite{DBLP:journals/tifs/ZhangXXLCFZ22}. Recently, Re-ID has gained increasing attention from academia and industry for its potential in video surveillance applications~\cite{zheng2016person,ye2020deep,DBLP:journals/tcsv/LengYT20,DBLP:journals/tifs/ZhuJYZSZ18}, which involves matching images of the same person captured by different cameras with non-overlapping camera views. However, the main challenge of Re-ID is the variations in body pose, viewing angle, illumination, image resolution, occlusion, background, etc., across different cameras~\cite{DBLP:journals/pami/LiZG20,chen2021occlude,DBLP:journals/tcsv/QiWHSG20,DBLP:journals/tifs/MaJZTP20}. Generally, person Re-ID can be seen as a special case of the image retrieval problem, where the objective is to quickly and accurately retrieve images from a large-scale gallery set that match a given query image.

\begin{figure}%[!h]
\centering
\subfigure[Sample-instance space]{
\includegraphics[width=4.10cm]{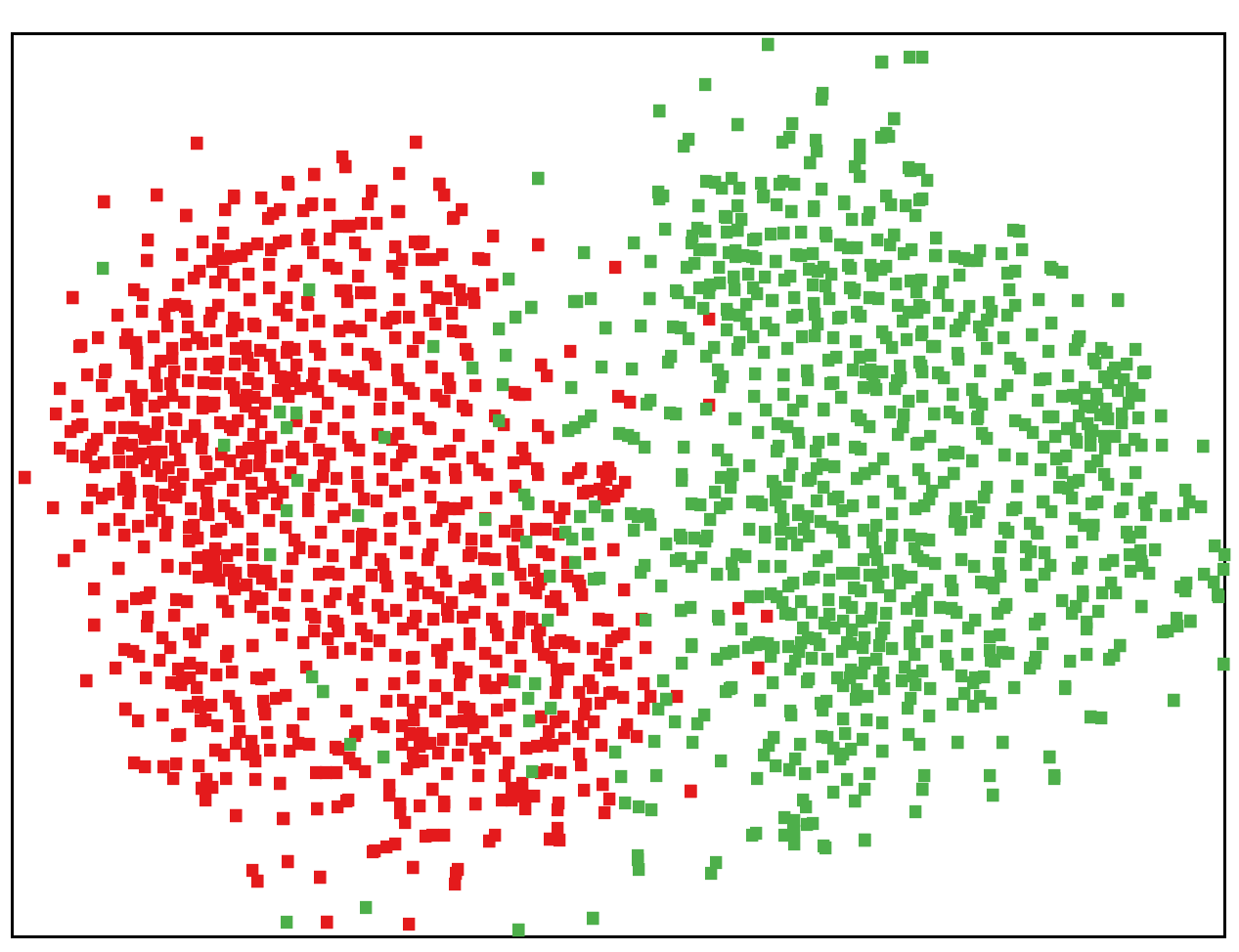}
}
\subfigure[Sample-pair space]{
\includegraphics[width=4.14cm]{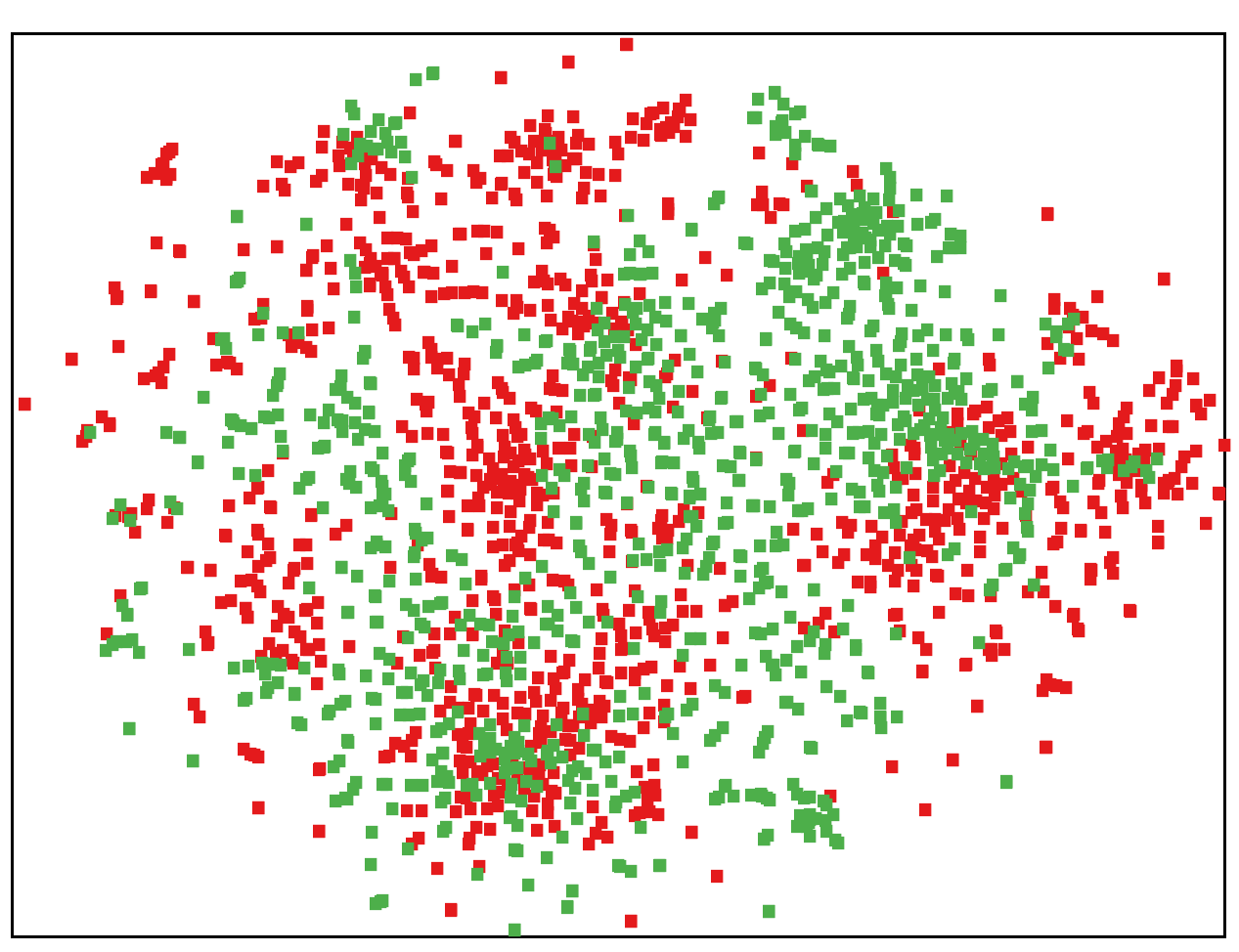}
}
\caption{Visualization of features in sample-instance and sample-pair spaces by t-SNE~\cite{van2008visualizing}. In Figure a), each dot represents a sample-instance feature, while in Figure b), each dot represents a sample-pair feature. For example, if there are two sample-instance features with $d$-dimension, \ie, the features of two images are extracted by ResNet, say $x=[x_1, \cdots, x_d] \in \mathbb{R}^{d}$ and $y=[y_1, \cdots, y_d] \in \mathbb{R}^{d}$, then the corresponding sample-pair feature can be represented as $xy=[(x_1-y_1)^2, \cdots, (x_d-y_d)^2]\in \mathbb{R}^{d}$. Both sample-instance and sample-pair features are normalized using $L_2$ normalization. Different colors denote different domains or datasets.}
\label{fig04}
  \vspace{-15pt}
\end{figure}

Deep learning has enabled excellent performance for the person re-identification task in supervised settings \cite{DBLP:journals/tmm/ZhaoLZZWM20,DBLP:journals/tmm/WeiZY0019,DBLP:conf/cvpr/WuZGL19,DBLP:conf/cvpr/LiZG18,DBLP:conf/cvpr/ZhengYY00K19,DBLP:journals/tomccap/QiWHSG21}. However, the performance drops significantly when these models are applied to new domains due to the data-distribution shift between the source domain and the unseen target domain. Collecting and labeling data is expensive and time-consuming, making existing supervised methods unsuitable for real-world applications. While some unsupervised domain adaptation methods have been developed to mitigate the labeling task \cite{DBLP:journals/tmm/YangYLJXYGHG21,DBLP:conf/cvpr/ZhaiLYSCJ020,DBLP:conf/iccv/WuZL19,DBLP:conf/eccv/ChenLL020,DBLP:conf/iccv/QiWHZSG19,DBLP:journals/tifs/LiCTYQ21,DBLP:journals/tifs/KhatunDSF21,lin2022ensemble}, they still require data collection and model re-training for new scenarios. Therefore, the domain generalization scenario remains a challenging task in person re-identification.

To overcome the aforementioned challenge, Domain Generalization (DG) methods have been developed to learn a model in the source domains and test it in the unseen target domain~\cite{zhou2021domain}. In the field of person re-identification, several DG methods have been proposed to obtain a robust model in the target domain. For instance, in QAConv~\cite{DBLP:conf/eccv/LiaoS20}, query-adaptive convolution kernels are utilized to achieve local matching by treating image matching as finding local correspondences in feature maps. In~\cite{DBLP:conf/cvpr/ZhaoZYLLLS21}, meta-learning is employed to mimic the train-test process of domain generalization for learning robust models. An effective voting-based mixture mechanism is leveraged in RaMoE~\cite{DBLP:conf/cvpr/DaiLLTD21} and META~\cite{xu2022mimic}, which dynamically utilizes the diverse characteristics of source domains to improve model's generalization. ACL~\cite{zhang2022adaptive} captures both domain-specific and domain-invariant features in a common feature space while exploring the relations across different domains. Unlike these existing methods, we propose a new perspective to address this issue.

% Domain generalization (DG) methods aim to address the aforementioned problem by learning a model in the source domains and testing the model in the unseen domain~\cite{zhou2021domain}. In the person Re-ID community, several DG methods have been developed to obtain a robust model in the unseen target domain. For example, QAConv~\cite{DBLP:conf/eccv/LiaoS20} treats image matching as finding local correspondences in feature maps and constructs query-adaptive convolution kernels on the fly to achieve local matching. In~\cite{DBLP:conf/cvpr/ZhaoZYLLLS21}, a meta-learning strategy is introduced to simulate the train-test process of domain generalization for learning more generalizable models. RaMoE~\cite{DBLP:conf/cvpr/DaiLLTD21} and META~\cite{xu2022mimic} adopt an effective voting-based mixture mechanism to dynamically leverage the diverse characteristics of source domains to improve the generalization ability of the model. ACL~\cite{zhang2022adaptive} maintains a common feature space for capturing both the domain-invariant and the domain-specific features while dynamically mining the relations across different domains.
% Different from these methods, we propose a new perspective to solve the issue.

In this paper, we concentrate on the cross-domain (\ie, domain generalization) person Re-ID as in \cite{DBLP:conf/cvpr/ZhaoZYLLLS21,qi2022label}, where several source domains are accessible in the training stage, meanwhile the data-distribution of the test (\ie, targe) domain is unknown. We define two kinds of features, called sample-instance features and sample-pair features, whose specific definitions are elaborated in the illustration of Fig. \ref{fig04}. Note that the sample-pair feature is not a pair of features, but a new feature constructed from the sampled pair of features. We find that the domain-shift across different domains is smaller in the sample-pair space than in the sample-instance space, as illustrated in Fig. \ref{fig04}. In this experiment, we randomly select $500$ samples from two datasets (Market1501 \cite{DBLP:conf/iccv/ZhengSTWWT15} and MSMT17 \cite{wei2018person}) respectively, and we then use the ResNet-50 \cite{DBLP:conf/cvpr/HeZRS16} pre-trained on ImageNet \cite{DBLP:conf/cvpr/DengDSLL009} to extract their features. For the sample-instance in Fig. \ref{fig04} (a), we directly visualize these sample-instance features by t-SNE \cite{van2008visualizing}. For the sample-pair in Fig. \ref{fig04} (b), we first randomly generate 500 sample-pairs for each domain and then produce sample-pair features using the subtract operation and the dot-squared operation. To ensure an equitable comparison, we perform the $L_2$ normalization for both sample-instance and sample-pair features. As can be seen from the figure, the sample-instance features from different domains are separated from each other, whereas the sample-pair features from different domains overlap with each other.

Building on the observation that sample-pair features are less affected by domain-shift than sample-instance features, we present the Generalizable Metric Network (GMN) to improve the model's discrimination and generalization. By leveraging the sample-similarity in the sample-pair space, our method effectively reduces the impact of the domain-shift across different domains. Specifically, we incorporate a Metric Network (M-Net) after the main backbone and train it using sample-pair features. The M-Net is then used to obtain the sample-pair similarity in the test stage. To further improve the generalization capability of the GMN, we introduce Dropout-based Perturbation (DP) at the feature-level, producing diverse sample-pair features that can sufficiently train the M-Net. Additionally, we design a Pair-Identity Center (PIC) loss to further boost the model's discrimination by pulling all pairs with the same pair-identity closer to its pair-center in each batch. After conducting experiments on various commonly used person Re-ID datasets, our proposed method has been demonstrated to be effective in improving the performance of the Re-ID task. Furthermore, detailed analysis has shown that our method has a significant advantage over existing methods.

In summary, our main contributions can be listed as:
  \begin{itemize}
    \item We develop a novel Generalizable Metric Network (GMN) for DG Re-ID, which can reduce the impact of the domain-shift based on the sample-pair space to improve the generalization ability of the model. 
    \item To further enhance the model's generalization and discrimination, we introduce Dropout-based Perturbation (DP) and Pair-Identity Center (PIC) loss.
    \item Our GMN achieves state-of-the-art accuracy on several standard benchmark datasets, demonstrating its superiority over existing state-of-the-art methods. 
  \end{itemize}

The structure of this paper is outlined as follows: Section \ref{s-related} provides a literature review on relevant research. In Section \ref{s-framework}, we introduce our proposed approach. Section \ref{s-experiment} presents the experimental results and analysis. Lastly, we summarize this paper in Section \ref{s-conclusion}.

% The rest of this paper is organized as follows.
% We review some related work in Section \ref{s-related}.
% The proposed method is introduced in Section \ref{s-framework}.
% Experimental results and analysis are presented in Section \ref{s-experiment},
% and Section \ref{s-conclusion} is conclusion.

\section{Related work}\label{s-related}
In this section, we will review the related approaches to our method, including generalizable person Re-ID, domain generalization and metric network (M-Net) based person Re-ID. The following part presents a detailed investigation.

\subsection{Generalizable Person Re-ID}
% Person Re-ID methods have achieved a great success in computer vision in recent years. For example, in \cite{DBLP:journals/tip/ZhangGDWZC21}, a novel Deep High-Resolution Pseudo-Siamese Framework (PS-HRNet) is introduced to solve the matching problem of person with the same identity but different resolutions captured by different cameras, which can alleviate the difference of feature distributions between low-resolution images and high-resolution images. In order to address the cross-illumination person Re-ID task, Zhang~\etal~\cite{9761930} develop a novel Illumination Estimation and Restoring framework (IER), which can effectively reduce the disparities between training and test images. Differently, the goal of domain generalizable person re-identification is to learn a robust model in the source domain that can directly perform well in the target domain without additional training. The existing methods mainly include network normalization, meta-learning and domain alignment. 
In recent years, person re-identification has become a significant research area in computer vision due to its practical applications. For example, Zhang~\etal propose a novel framework called Deep High-Resolution Pseudo-Siamese Framework (PS-HRNet) in \cite{DBLP:journals/tip/ZhangGDWZC21} to address the challenge of matching persons with the same identity captured at different resolutions. The PS-HRNet framework reduces the differences in feature distributions between low-resolution and high-resolution images. In another work, Zhang~\etal develop an Illumination Estimation and Restoring framework (IER) \cite{9761930} to address the task of cross-illumination person re-identification. The IER framework effectively reduces the disparities between training and test images.
However, the objective of domain generalizable person re-identification is to learn a model that can perform well in target domains without additional training, by reducing the domain gap between the source and target domains. Existing methods for domain generalizable person re-identification mainly involve techniques such as network normalization, meta-learning, and domain alignment.

The area of network normalization methods focuses on integrating Batch Normalization (BN) and Instance Normalization (IN) effectively for domain generalizable person re-identification \cite{DBLP:conf/nips/EomH19,DBLP:conf/bmvc/JiaRH19,DBLP:conf/cvpr/JinLZ0Z20,xu2022mimic}. For example, Jia~\etal~\cite{DBLP:conf/bmvc/JiaRH19} add IN in specific layers to remove shifts in style and content of different domains in Re-ID. However, this method can eliminate some important discriminative information. To overcome this, Jin~\etal~\cite{DBLP:conf/cvpr/JinLZ0Z20} propose a method for extracting identity-relevant features from removed information, adjusting and restoring them using a style normalization module, and reintroducing the features to the network to ensure high discrimination.
Moreover, in poor generalization scenarios, Choi~\etal~\cite{DBLP:conf/cvpr/ChoiKJPK21} explored using batch instance normalization to address overfitting.
In addition, Meta-learning~\cite{DBLP:conf/icpr/LinCW20,DBLP:conf/cvpr/SongYSXH19,DBLP:conf/cvpr/ZhaoZYLLLS21,DBLP:conf/cvpr/DaiLLTD21} is another approach used in network normalization methods. For instance, Song~\etal~\cite{DBLP:conf/cvpr/SongYSXH19} develop a weight network for the class-specific classifier, which can map domain-invariant information, and utilize meta-learning to enhance generalization performance on unseen datasets.
In~\cite{DBLP:conf/cvpr/ZhaoZYLLLS21}, the proposed method is a meta-learning framework based on memory, which aims to simulate the train-test process of domain generalization during training. The method incorporates a meta batch normalization layer to diversify the meta-test feature.

Domain alignment techniques aim to tackle the issue of dissimilar data distribution among various domains by mapping all samples to the same feature space~\cite{DBLP:conf/aaai/ChenDLZX0J21,DBLP:conf/eccv/LuoSZ20,DBLP:conf/wacv/YuanCCYRW020,DBLP:conf/eccv/ZhuangWXZZWAT20,DBLP:conf/eccv/LiaoS20}. Dual distribution alignment networks have been proposed by Chen \etal~\cite{DBLP:conf/aaai/ChenDLZX0J21} and Zhang \etal~\cite{zhang2022adaptive} to align the distributions of multiple source domains and achieve domain-invariant feature space. Yuan \etal~\cite{DBLP:conf/wacv/YuanCCYRW020} extracted identity-related characteristics by an adversarial framework to learn domain-invariant features from difficult variations, incorporating information such as video timestamp and camera serial number.
QAConv~\cite{DBLP:conf/eccv/LiaoS20} addresses image matching by finding local correspondences in feature maps and constructing query-adaptive convolution kernels for local matching. %Besides, Liao \etal~\cite{liao2021graph} introduce hard example mining during data sampling by constructing a nearest neighbor relationship graph for all classes, providing informative and challenging examples for learning.

Moreover, a novel Meta Distribution Alignment (MDA) method~\cite{ni2022meta} is proposed to enable them to share similar distribution in a test-time-training fashion.
Ni \etal~\cite{ni2023part} propose a pure Transformer model (termed Part-aware Transformer) for DG-ReID by designing a proxy task to mine local visual information shared by different IDs.
Jiao \etal~\cite{jiao2022dynamically} develop a new normalization scheme called Dynamically Transformed Instance Normalization (DTIN) to alleviate the drawback of IN (\ie, the limitation of eliminating discriminative patterns). 
Pu \etal~\cite{pu2021lifelong,pu2023memorizing} design an Adaptive Knowledge Accumulation (AKA) framework that is endowed with two crucial abilities: knowledge representation and knowledge operation~\cite{zheng2022faster}. Different from these methods, our method explores the generalization capability in ReID based on the metric network.

In this paper, we address the generalizable person Re-ID task using a metric network approach. Our proposed method has two main objectives: to explore the similarity in the sample-pair space, and to mitigate the data-distribution discrepancy across different domains, while simultaneously learning a domain-invariant metric network.

\subsection{Domain Generalization}
There has been a surge of interest in developing methods to tackle the domain generalization problem in classification and semantic segmentation tasks in recent years~\cite{DBLP:conf/cvpr/NamLPYY21,wang2022feature,DBLP:conf/iccv/YueZZSKG19,DBLP:conf/cvpr/CarlucciDBCT19,DBLP:conf/nips/BalajiSC18,DBLP:conf/iccv/LiZYLSH19,DBLP:conf/eccv/LiTGLLZT18,DBLP:journals/pr/ZhangQSG22,lingwal2023semantic}. Many of these approaches aim to mitigate the distribution discrepancy between different domains by mapping data into a common space, following the domain adaptation paradigm~\cite{DBLP:conf/eccv/LiTGLLZT18,DBLP:conf/nips/ZhaoGLFT20,DBLP:conf/icml/MuandetBS13,DBLP:journals/pr/RahmanFBS20,DBLP:conf/cvpr/LiPWK18,DBLP:conf/cvpr/GongLCG19,DBLP:conf/wacv/RahmanFBS19}. For instance, Muandet \etal~\cite{DBLP:conf/icml/MuandetBS13} propose a kernel-based optimization algorithm to learn domain-invariant features; however, it does not ensure the consistency of conditional distribution. To overcome this limitation, Zhao \etal~\cite{DBLP:conf/nips/ZhaoGLFT20} introduce an entropy regularization term to measure the dependence between the learned features and class labels, which promotes conditional invariance of the features and enables the classifier to perform well on features from different domains.

Furthermore, this method~\cite{DBLP:conf/cvpr/GongLCG19} employs CycleGAN~\cite{DBLP:conf/iccv/ZhuPIE17} to generate novel image styles that are absent in the training dataset, thus closing the gap between the source and target domains to enhance model generalization. Li~\etal~\cite{DBLP:conf/cvpr/LiPWK18} leverage an adversarial autoencoder learning framework to acquire a universal latent feature representation in the hidden layer by aligning source domains using Maximum Mean Discrepancy. They then match this aligned distribution to an arbitrary prior distribution via adversarial feature learning to improve the generalization of the hidden layer feature to other unknown domains. Rahman~\etal~\cite{DBLP:journals/pr/RahmanFBS20} incorporate a correlation alignment module and adversarial learning to create a more domain-agnostic model that can effectively reduce domain discrepancy. Moreover, in addition to performing adversarial learning at the domain level, Li~\etal~\cite{DBLP:conf/eccv/LiTGLLZT18} also execute domain adversarial tasks at the category level by aligning samples of each class from different domains, achieving great generalization.

Particularly, our method utilizes a metric network to further learn the domain-invariant evaluator in the sample-pair space, which has not been investigated in existing works. Moreover, since person Re-ID is a metric task (\ie, the goal is to identify whether two images show the same person or different persons, which is different from the classification task), it is also crucial to accurately mine the similarities between samples.

\subsection{The M-Net based Person Re-ID}
In the early days of deep learning~\cite{aversa2020deep}, several methods employed the two-stream metric structure to obtain the similarity metric of two samples in the supervised person Re-ID task~\cite{subramaniam2016deep,zhang2016semantics,li2014deepreid,ahmed2015improved}. For instance, Li~\etal~\cite{li2014deepreid} proposed a novel filter pairing neural network to jointly handle misalignment, photometric and geometric transforms, occlusions and background clutter. Ahmed~\etal~\cite{ahmed2015improved} computed cross-input neighborhood differences to capture local relationships between the two input images based on mid-level features from each input image. However, these methods require a pair of images for evaluation, which resulted in feeding an image into the whole network repeatedly. In recent years, most existing person Re-ID methods directly train a robust backbone network and extract sample features for evaluation.

Unlike previous M-Net based person Re-ID methods, our method involves developing an M-Net based on sample-pair features constructed using sample-instance features from the main backbone, specifically designed for the cross-domain person Re-ID task. This is a novel method that has not been explored in prior works. Furthermore, our method utilizes a single path, in contrast to the two-stream structure used in previous works such as~\cite{subramaniam2016deep,zhang2016semantics,li2014deepreid,ahmed2015improved}.

\section{The proposed method}\label{s-framework}

\begin{figure*}[t]
\centering
\includegraphics[width=16cm]{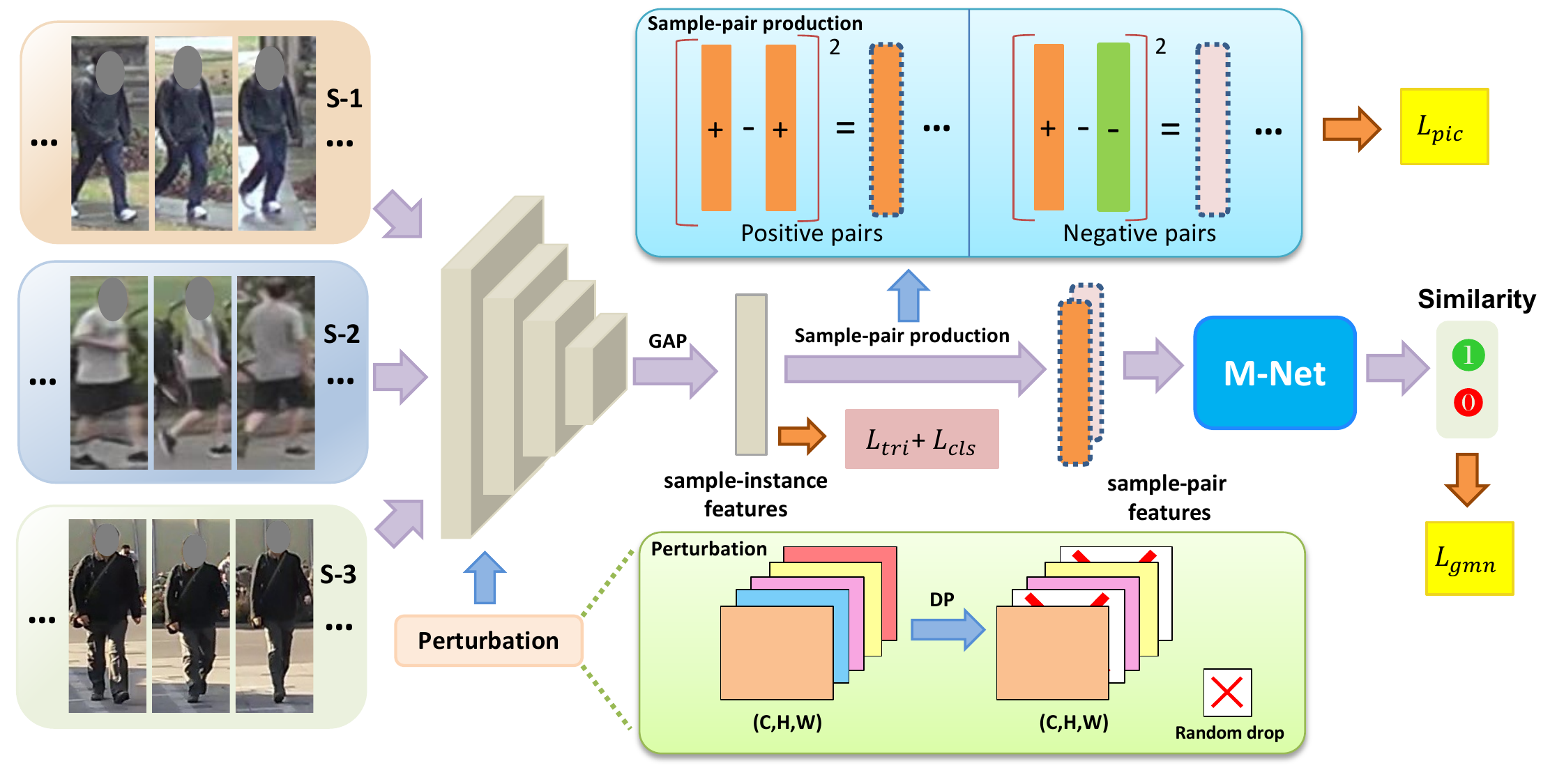}
\caption{An illustration of the proposed Generalizable Metric Network (GMN). Here we take three source domains as an example. As seen in this figure, our method yields sample-pair features based the main (backbone) network, which are leveraged to train the metric network (M-Net). Besides, the dropout-based perturbation (DP) and the pair-identity center loss ($\mathrm{L}_{pic}$) are introduced to improve the model's generalization and discrimination. It is worth noting that both backbone and M-Net are used during the test process.}
\label{fig03}
  \vspace{-15pt}
%\vspace*{-20pt}% µ÷Õû¼ä¾à
\end{figure*}

In this paper, to enhance the generalization capacity of the model to the unseen domain, we propose a new Generalizable Metric Network (GMN) for generalizable multi-source person re-identification (Re-ID), which consists of the backbone, the metric network (M-Net), the dropout-based perturbation (DP), the pair-identity center (PIC) loss and the typical Re-ID loss. Particularly, the proposed GMN is designed to alleviate the impact of the domain-shift across different domains. The overview of our method is displayed in Fig.~\ref{fig03}. In the next part, we will introduce the proposed method in detail.

\subsection{Generalizable Metric Network} \label{LDL-1}
In most existing generalizable person Re-ID methods, the focus is on obtaining a generalizable model that can extract domain-invariant or robust features for unseen domains. However, in this paper, we observe that the domain gap is smaller in the sample-pair feature space compared to the sample-instance feature space, as shown in Fig.~\ref{fig04}. Therefore, we propose a novel approach that utilizes a metric network (M-Net) for inference in the test stage, which differs from conventional person Re-ID methods that use features to conduct the metric. To train the M-Net, we first add it after the backbone, and then produce positive sample-pairs and negative sample-pairs based on sample-instance features extracted by the backbone in a batch. Let $f^{(si)}_p$ and $f^{(si)}_q$ denote the sample-instance features of the $p$-th and $q$-th samples, respectively, and we can obtain the $k$-th sample-pair feature as:
\begin{equation}
  \begin{aligned}
  f^{(sp)}_k = (f^{(si)}_p-f^{(si)}_q)\odot(f^{(si)}_p-f^{(si)}_q) ,
  \end{aligned}
  \label{eq01}
  \end{equation}
  where $\odot$ is the element-wise multiplication operation. If $f^{(si)}_p$ and $f^{(si)}_q$ have the same identity, $f^{(sp)}_k$ is considered as a positive sample-pair feature, otherwise it is considered as a negative sample-pair feature. We use $f^{(sp)}_k$ as the input to train the metric network. As the input to the metric network only has two classes (positive and negative), we use a binary classification loss to train the metric network, which is given by:
\begin{equation}
  \begin{aligned}
  \mathrm{L}_{gmn} =
\begin{cases}
  & -\log(P(1|f^{(sp)}_{k})) \text{ if } f^{(sp)}~is~positive;\\
  & -\log(P(0|f^{(sp)}_{k})) \text{ if } f^{(sp)}~is~negative,
\end{cases}
  \end{aligned}
  \label{eq02}
  \end{equation}
  where $P(1|f^{(sp)}_k)$ and $P(0|f^{(sp)}_k)$ are the output of the metric network for the input $f^{(sp)}_k$. We use $\mathrm{L}_{gmn}$ and the typical person Re-ID losses (\ie, triplet loss and classification loss) together to train the main network and the metric network.

 In the test stage, we first extract sample-instance features ($F^{(si)}{p}\in \mathbb{R}^{d\times N_p}$ and $F^{(si)}{g}\in \mathbb{R}^{d\times N_g}$, where $d$ is the feature dimension and $N_p$ and $N_g$ are the numbers of samples in the probe and gallery sets) from the backbone network. We then use Eq.~\ref{eq01} to form sample-pair features $F^{(sp)} \in \mathbb{R}^{d\times N_pN_g}$ and feed them into the metric network to produce a similarity matrix $S\in \mathbb{R}^{N_p\times N_g}$, which is used for evaluating performance. The similarity between the $i$-th and $j$-th samples can be represented as:

  \begin{equation}
  \begin{aligned}
  S_{ij} = \frac{exp(P(1|f^{(sp)}_{ij}))}{exp(P(1|f^{(sp)}_{ij}))+exp(P(0|f^{(sp)}_{ij}))}.
  \end{aligned}
  \label{eq01}
  \end{equation}

  \textit{Remark:} In this paper, we use two fully connected layers and ReLU to implement the metric network, which contains very few parameters when compared to the backbone network. Therefore, using the metric network in our method does not significantly increase the computation in the training stage. Additionally, in the test stage, we employ the pre-extracting-feature scheme to reduce the inference time, which can achieve similar inference times as the feature-metric evaluation. We will validate this in the experimental section.

% When finishing a epoch, we update $\mathbf{M_L}$ as:
%   \begin{equation}
%   \begin{aligned}
%   \mathbf{M_L} = \mathbf{M_T}
%   \end{aligned}
%   \label{eq02}
%   \end{equation}

% \begin{algorithm}[ht]
% \caption{\small{The forward process of mix-normalization (DMN)}}~\label{alg1}%\newline % Ëã·¨µÄÃû×Ö
% \begin{algorithmic}[1]
% \STATE {\bf Input:} % Ëã·¨µÄÊäÈë£¬ \hspace*{0.02in} ÓÃÀ´¿ØÖÆÎ»ÖÃ£¬Í¬Ê±ÀûÓÃ \\ ½øÐÐ»»ÐÐ
% Feature $f$ of all $D\times P$ samples in a batch and the initial domain set $\mathbf{S}$ including all $D$ source domains.\\
% \STATE {\bf Output:} The normalized feature $\hat{f}$. \\
% %\begin{algorithmic}[1]
% \STATE Randomly produce the number of mixing domains as $C$.
% \WHILE{$|\mathbf{S}|\neq 0$}
% \IF {$|\mathbf{S}|=<C$}
% \STATE Set $C$ as $|\mathbf{S}|$.
% \ENDIF
% %\STATE \textcolor{blue}{// Fix feature extractor $\theta_{F}$, and update discriminator $\theta_{D}$.}\\
% \STATE Randomly select $C$ domains $\phi$ from the $\mathbf{S}$.\\
% \STATE Compute the statistics (\ie, mean and variance) for the selected $C$ domains as Eqs.~\ref{eq05} and ~\ref{eq06}.\\
% %\STATE \textcolor{blue}{// Fix discriminator $\theta_{D}$, and update feature extractor $\theta_{F}$.}\\
% \STATE Utilize the statistics to normalize all samples from the selected $C$ domains as Eq.~\ref{eq04}.
% %\STATE Update the statistics for evaluation as Eqs.
% \STATE Remove the selected $C$ domains from the $\mathbf{S}$.\\
% \ENDWHILE
% \end{algorithmic}
% \label{al01}
% \end{algorithm}

\subsection{Improving the Robustness of the GMN}\label{LDL-2}
To enhance the robustness of the GMN, we aim to increase the diversity of sample-pairs at the feature-level. To achieve this, we leverage a dropout-based perturbation (DP) method. This involves applying dropout to the channels of the feature maps, resulting in the random dropping of some channels to perturb the feature maps. Since the DP module does not destroy the spatial structure information, it can preserve the semantic information of each image. We insert the DP module into the middle layers of the backbone network, as the deeper layers of neural networks focus on semantic information. This helps maintain the semantic information of the image while increasing the diversity of sample-pairs.

Additionally, we propose a pair-identity center (PIC) loss to further improve the discrimination of the GMN. Since sample-pairs are produced for the M-Net, we use the PIC loss to make all sample-pair features with the same pair-identity close, which can produce discriminative sample-pair features for the M-Net. To achieve this goal, we first define the positive sample-pair features as:

\begin{equation}
  \begin{aligned}
  f^{(sp)}_{1+} = (f^{(si)}_p-f^{(si)}_{p+})\odot(f^{(si)}_p-f^{(si)}_{p+}),\\
    f^{(sp)}_{2+} = (f^{(si)}_p-\bar{f}^{(si)}_{p})\odot(f^{(si)}_p-\bar{f}^{(si)}_{p}),\\
      f^{(sp)}_{3+} = (\bar{f}^{(si)}_p-f^{(si)}_{p+})\odot(\bar{f}^{(si)}_p-f^{(si)}_{p+}),
  \end{aligned}
  \label{eq04}
  \end{equation}
where $\bar{f}^{(si)}_{p}$ is the averaged feature of all samples with the same identity with $f^{(si)}_{p}$ in a batch, and $f^{(si)}_{p+}$ is the same identity as $f^{(si)}_{p}$. Thus, we can obtain the pair-identity loss for the positive sample-pairs as:

  \begin{equation}
  \begin{aligned}
  \mathrm{L}_{pic+} = \frac{1}{3}\sum_{i,j\in\{1,2,3\}}^{i\ne j}\left \| f^{(sp)}_{i+} - f^{(sp)}_{j+} \right \|.
  \end{aligned}
  \label{eq05}
  \end{equation}

  Similarly, for the negative sample-pair, we can obtain the sample-pair features as:
  \begin{equation}
  \begin{aligned}
  f^{(sp)}_{1-} = (f^{(si)}_p-f^{(si)}_{q})\odot(f^{(si)}_p-f^{(si)}_{q}),\\
    f^{(sp)}_{2-} = (f^{(si)}_p-\bar{f}^{(si)}_{q})\odot(f^{(si)}_p-\bar{f}^{(si)}_{q}),\\
      f^{(sp)}_{3-} = (\bar{f}^{(si)}_p-f^{(si)}_{q})\odot(\bar{f}^{(si)}_p-f^{(si)}_{q}),\\
       f^{(sp)}_{4-} = (\bar{f}^{(si)}_p-\bar{f}^{(si)}_{q})\odot(\bar{f}^{(si)}_p-\bar{f}^{(si)}_{q}).
  \end{aligned}
  \label{eq06}
  \end{equation}
Thus, we can compute the pair-identity loss for the negative sample-pairs as:

  \begin{equation}
  \begin{aligned}
  \mathrm{L}_{pic-} = \frac{1}{6}\sum_{i,j\in\{1,2,3,4\}}^{i\ne j}\left \| f^{(sp)}_{i-} - f^{(sp)}_{j-} \right \|.
  \end{aligned}
  \label{eq07}
  \end{equation}
Therefore, the whole PIC loss can be defined by:
  \begin{equation}
  \begin{aligned}
  \mathrm{L}_{pic} =  \mathrm{L}_{pic+} +  \mathrm{L}_{pic-}.
  \end{aligned}
  \label{eq08}
  \end{equation}

\textit{Remark:} The PIC loss proposed in this paper differs from the typical center loss~\cite{DBLP:conf/eccv/WenZL016} in two aspects. First, it aims to bring positive sample-pair features with the same pair-identity closer, which is more suitable for the metric-based person Re-ID task that evaluates the similarity between samples. Second, our PIC loss is computed in a batch. This is because the negative-pair-identity is infinite, making it impossible to define a center for each negative-pair-identity on the entire dataset.
\subsection{The Training Process}
During training, the model is trained using three different loss functions: cross-entropy loss ($\mathrm{L}_{cls}$), triplet loss ($\mathrm{L}_{tri}$) with hard mining sampling~\cite{ulyanov2016instance}, and the loss of the M-Net ($\mathrm{L}_{gmn}$), and pair-identity center loss ($\mathrm{L}_{pic}$). The cross-entropy loss and the triplet loss are commonly used in person Re-ID research~\cite{DBLP:journals/tmm/LuoJGLLLG20,DBLP:conf/iccv/FuWWZSUH19}. The overall loss function used for training the model can be expressed as follows:

\begin{equation}
\mathrm{L}_{overall}=\mathrm{L}_{cls}+\mathrm{L}_{tri}+\mathrm{L}_{gmn}+\lambda \mathrm{L}_{pic},
\label{eq09}
\end{equation}
where $\lambda$ is a hyper-parameter that balances the basic loss functions and the pair-identity center (PIC) loss. The generalizable metric network is implemented using Alg.~\ref{al01}.

\textit{Remark:} To improve the robustness of the network for the person Re-ID task, the dropout-based perturbation (DP) module is not activated at the beginning of the training process, as the initialized network is not robust enough for such samples. As shown in  Alg.~\ref{al01}, the DP module is activated after a certain number of epochs when the model is trained. The effectiveness of this method will be evaluated in the experimental section.

% \begin{algorithm}[t]
% \caption{\small{The training process of the proposed GMN}}~\label{alg1}%\newline % Ëã·¨µÄÃû×Ö
% \begin{algorithmic}[1]
% \STATE {\bf Input}: Training examples $\mathrm{X}$ and labels $\mathrm{Y}$.
% \STATE {\bf Output}: The backbone ($\theta^{b}$) and M-Net ($\theta^{mn}$).
% \STATE {\bf procedure} % Ëã·¨µÄÊäÈë£¬ \hspace*{0.02in} ÓÃÀ´¿ØÖÆÎ»ÖÃ£¬Í¬Ê±ÀûÓÃ \\ ½øÐÐ»»ÐÐ
% %LDL %(Training examples $\mathrm{X}$, labels $\mathrm{Y}$) \\
% \STATE $\theta^{b} \leftarrow$ Initialize by ResNet-50 pre-trained on ImageNet.\\
% \STATE $\theta^{mn} \leftarrow$ Initialize by random parameters. \\
% \tcp{\bl{\scriptsize{The number of epochs is $T$.}}} 
% \tcp{\bl{\scriptsize{We activate DP at the $T^{'}$-th epoch.}}}
% \FOR{epoch $\in [1,...,T]$}
% \IF{epoch $>= T^{'}$}
% \STATE Activate the dropout-based on perturbation module.
% \ENDIF \\
% \tcp{\bl{\scriptsize{The number of iterations of each epoch is $N$.}}}

% \FOR{iteration $\in [1,...,N]$}
% \STATE Feed a batch into the backbone to obtain sample-instance features.
% \STATE Produce sample-pair features as Eq.~\ref{eq01}.
% \STATE Feed sample-pair features into M-Net.
% \STATE Compute the whole loss as Eq.~\ref{eq09}.
% \ENDFOR
% \ENDFOR
% \STATE {\bf end procedure}
% \end{algorithmic}
% \label{al01}
% \end{algorithm}

\begin{algorithm}[t]\label{al01}
    \SetAlgoLined %显示end
	\caption{\small{The training process of the proposed GMN}}%算法名字
	\KwIn{Training samples $\mathrm{X}$ and labels $\mathrm{Y}$.}%输入参数
	\KwOut{The backbone ($\theta^{b}$) and M-Net ($\theta^{mn}$).}%输出
	 $\theta^{b} \leftarrow$ Initialize by ResNet-50 pre-trained on ImageNet. \\ %\;用于换行
	 $\theta^{mn} \leftarrow$ Initialize by random parameters.\\
	 \tcp{\bl{\scriptsize{The number of epochs is $T$.}}} 
	\For{epoch $\in [1,...,T]$}{
	\tcp{\bl{\scriptsize{We activate DP at the $T^{'}$-th epoch.}}}
		\If{epoch $>= T^{'}$}{
			Activate the dropout-based on perturbation module.
		}
		\tcp{\bl{\scriptsize{The number of iterations in each epoch is $N$.}}}
		\For{iteration $\in [1,...,N]$}{
		 Feed a batch into the backbone to obtain sample-instance features.\\
		 Produce sample-pair features as Eq.~\ref{eq01}.\\
		 Feed sample-pair features into M-Net.\\
		 Compute the whole loss as Eq.~\ref{eq09}.
		}
	}
	return 

\end{algorithm}
\subsection{Explanation for GMN via Existing Theory}

In this section, we use the domain generalization error bound~\cite{albuquerque2019generalizing} to demonstrate the effectiveness of our method. Firstly, we review the domain generalization error bound, and then we analyze our method based on it.

\textbf{Theorem 1}~\cite{albuquerque2019generalizing,DBLP:conf/ijcai/0001LLOQ21} (Domain generalization error bound): Let $\gamma := min_{\pi \in \bigtriangleup_{M}} d_{\mathcal{H}}(\mathcal{P}_{X}^t, \sum_{i=1}^{M}\pi_{i}\mathcal{P}_{X}^i)$\footnote{$M$ is the number of source domains.} with minimizer $\pi^*$ being the distance of $P_{X}^t$ from the convex hull $\Lambda$. Let $P_{X}^* := \sum_{i=1}^{M}\pi_{i}^{*}P_{X}^{i}$ be the best approximator within $\Lambda$. Let $\rho := \sup_{\mathcal{P}_{X}^{'}, \mathcal{P}_{X}^{''}\in \Lambda}d_{\mathcal{H}}(\mathcal{P}_{X}^{'}, \mathcal{P}_{X}^{''})$ be the diameter of $\Lambda$. Then it holds that

\begin{equation}
\begin{aligned}
\epsilon^{t}(h)\leqslant \sum_{i=1}^{M}\pi_{i}^{*}\epsilon^{i}(h)+\frac{\gamma+\rho}{2}+\lambda_{\mathcal{H}}(\mathcal{P}_X^t, \mathcal{P}_X^*)),
\end{aligned}
\label{eq22}
\end{equation}
where $\lambda_{\mathcal{H}}(\mathcal{P}_X^t, \mathcal{P}_X^*)$ is the ideal joint risk across the target domain and the training domain ($P_X^*$) with the most similar distribution to the target domain.

In Theorem 1, the last item can be treated as a constant because it represents the ideal joint risk across the target domain and the training domain ($P_X^*$) with the most similar distribution to the target domain. Therefore, we primarily focus on analyzing the first item and the second item (which consist of $\gamma$ and $\rho$).

Firstly, the first item is the empirical risk error of all source domains. Unlike the typical classification task, the person Re-ID task aims to obtain a robust model to measure the similarity between two samples. The proposed PIC loss can effectively enhance the sample-pair discrimination, thereby reducing the empirical risk error in the training set.

Secondly, $\gamma$ represents the discrepancy between the combination of all training domains and the target domain. In the domain generalization person Re-ID setting, if the test domain is far from the training domain in terms of distribution, the model's generalization may be poor for all test samples. Particularly, our method utilizes a metric network to train and test in the sample-pair space where there is a small domain gap across different domains, as shown in Fig.~\ref{fig04}. Therefore, introducing the metric network can be beneficial in reducing the influence of the domain-shift between training and test sets and effectively mitigating the aforementioned risk.

Thirdly, $\rho$ indicates the maximum distance between different domains. In our method, the DP scheme preserves semantic information while not introducing additional information. This indicates that generating diverse style information in our method does not create a large domain gap between training samples. Additionally, we apply the DP module to the middle layer of the neural network (which mainly focuses on style information), helping to prevent the creation of a large $\rho$. Moreover, the domain gap is smaller in the sample-pair space when compared with the sample-instance space. In summary, our method has the advantage of reducing the generalization error bound from both the first and second items in Eq.~\ref{eq22}.

\subsection{Discussion}

We will clarify the distinctions between M-Net and~\cite{zheng2017discriminatively,chen2018deep} from the following four aspects. Firstly, M-Net generates sample-pair features at the feature level, which helps reduce computational costs. In contrast, prior works~\cite{zheng2017discriminatively} and~\cite{chen2018deep} generate sample-pair images at the image level, resulting in a disparity in the positions where sample-pair features are formed. Additionally, in terms of network architecture, both~\cite{zheng2017discriminatively} and~\cite{chen2018deep} are based on the framework of Siamese networks, whereas our approach deviates from this paradigm. Secondly, the approach in~\cite{chen2018deep} employs dropout on the difference vector, whereas we apply dropout on the feature maps. Our objective is to enhance the diversity of features to better address domain generalization tasks. Thirdly, in comparison to~\cite{zheng2017discriminatively} and~\cite{chen2018deep}, although M-Net utilizes the same loss function, we further introduce the pair-identity center loss to generate more discriminative features. Fourthly, while~\cite{zheng2017discriminatively} and~\cite{chen2018deep} have already adopted metric learning approaches to address the problem of person ReID in supervised scenarios, to the best of our knowledge, we are the first to introduce metric network for tackling DG-ReID. Our motivation is to address the challenges of generalization in this context.

Moreover, in~\cite{duin2012dissimilarity} and~\cite{yang2018dissimilarity}, the dissimilarity representation is a vector that combines the Euclidean distances of an instance to all other instances in the set. These dissimilarity representations form the dissimilarity space of the sample set. In contrast, our sample-pair space is not related to the Euclidean distance between sample pairs. We have defined new sample-pair features that can better assist the metric network in learning domain-invariant information.

Besides, compared to these methods~\cite{wu2016personnet,wu2018and,wu2019cross}, they use the siamese network architecture via requiring a pair of images as input. For example, when we compute the similarity between the $i$-th image and the $j$-th image $\&$ the $k$-th image, the $i$-th image is required to input the backbone two times. Different from these methods, our method uses a single network as the backbone and form the sample-pair feature in the feature space, which can effectively reduce the computation cost during training and testing. Moreover, these methods are developed to solve the supervised ReID task, while our method aims to deal with the DG-ReID task.

The dropout-based method we introduced differs from the approach in~\cite{hou2019weighted} in two aspects. Firstly, the dropout mechanism varies. In~\cite{hou2019weighted}, dropout in the convolutional layer involves generating a binary mask to select corresponding channels, whereas our approach, as depicted in Fig.~\ref{fig03} in the paper, involves randomly masking an entire portion of a channel in the feature maps. Secondly, the purpose of dropout is distinct. In~\cite{hou2019weighted}, dropout aims to address overfitting issues, while our objective is to enable the metric network to learn more diverse feature representations. Furthermore, the primary motivation behind introducing DP in GMN is to address the generalization challenges in the DG-ReID task more effectively.

\section{Experiments}\label{s-experiment}
\renewcommand{\cmidrulesep}{0mm} 
\setlength{\aboverulesep}{0mm} 
\setlength{\belowrulesep}{0mm} 
\setlength{\abovetopsep}{0cm}  
\setlength{\belowbottomsep}{0cm}

\newcommand{\PreserveBackslash}[1]{\let\temp=\\#1\let\\=\temp}
\newcolumntype{C}[1]{>{\PreserveBackslash\centering}p{#1}}
\newcolumntype{R}[1]{>{\PreserveBackslash\raggedleft}p{#1}}
\newcolumntype{L}[1]{>{\PreserveBackslash\raggedright}p{#1}}

\begin{table*}[htbp]
  \centering
  \caption{Comparison with SOTA methods on Market1501 (M), MSMT17 (MS) and CUHK03 (C3). The top and the bottom are results using the training set and all images on each dataset, respectively. Note that the \textbf{bold} is the best result and {\dag} indicates evaluation results based on the code released by the authors.}
    \begin{tabular}{c|c|c|cc|cc|cc|cc}
    \toprule
    \multirow{2}[1]{*}{Setting} & \multirow{2}[1]{*}{Method} & \multirow{2}[1]{*}{~~Backbone~~} & \multicolumn{2}{c|}{~~~M+MS+CS$\rightarrow$C3~~~} & \multicolumn{2}{c|}{~~~M+CS+C3$\rightarrow$MS~~~} & \multicolumn{2}{c|}{~~~MS+CS+C3$\rightarrow$M~~~} & \multicolumn{2}{c}{Average} \\
\cmidrule{4-11}          &       &       & mAP   & R1  & mAP   & R1 & mAP   & R1  & ~~mAP~~   & ~~R1~~  \\
    \midrule
    \multirow{7}[2]{*}{Training Sets} & SNR~\cite{DBLP:conf/cvpr/JinLZ0Z20}  & ResNet50-IN & 8.9  & 8.9   & 6.8   & 19.9  & 34.6  & 62.7  & 16.8  & 30.5 \\
          & QAConv$_{50}$~\cite{DBLP:conf/eccv/LiaoS20} & ResNet50 & 25.4 & 24.8  & 16.4  & 45.3  & 63.1  & 83.7  & 35.0  & 51.3 \\
          & M$^3$L~\cite{DBLP:conf/cvpr/ZhaoZYLLLS21}  & ResNet50-IBN & 34.2 & 34.4  & 16.7  & 37.5  & 61.5  & 82.3  & 37.5  & 51.4 \\
          & MetaBIN~\cite{DBLP:conf/cvpr/ChoiKJPK21}& ResNet50-BIN & 28.8 & 28.1  & 17.8  & 40.2  & 57.9  & 80.1  & 34.8  & 49.5 \\
          & META~\cite{xu2022mimic}  & ResNet50-DSBN & 36.3 & 35.1  & 22.5  & 49.9  & 67.5  & 86.1  & 42.1  & 57.0 \\
          & ACL~\cite{zhang2022adaptive}   & ResNet50-IBN & 41.2 & 41.8  & 20.4 & 45.9  & \textbf{74.3} & \textbf{89.3} & 45.3  & 59.0 \\
          & GMN(Ours)   & ResNet50-IBN  & \textbf{43.2} & \textbf{42.1} & \textbf{24.4} & \textbf{50.9} & 72.3  & 87.1  & \textbf{46.6} & \textbf{60.0} \\
    \midrule
    \multirow{9}[2]{*}{Full Images} & SNR~\cite{DBLP:conf/cvpr/JinLZ0Z20}  & ResNet50-IN & 17.5 & 17.1  & 7.7   & 22.0  & 52.4  & 77.8  & 25.9  & 39.0 \\
          & QAConv$_{50}$~\cite{DBLP:conf/eccv/LiaoS20} & ResNet50 & 32.9 & 33.3  & 17.6  & 46.6  & 66.5  & 85.0  & 39.0  & 55.0 \\
          & M$^3$L~\cite{DBLP:conf/cvpr/ZhaoZYLLLS21} & ResNet50-IBN & 35.7 & 36.5  & 17.4  & 38.6  & 62.4  & 82.7  & 38.5  & 52.6 \\
          & MetaBIN~\cite{DBLP:conf/cvpr/ChoiKJPK21} & ResNet50-BIN & 43.0 & 43.1  & 18.8  & 41.2  & 67.2  & 84.5  & 43.0  & 56.3 \\
          & META~\cite{xu2022mimic}  & ResNet50-DSBN & 47.1 & 46.2  & 24.4  & 52.1  & 76.5  & 90.5  & 49.3  & 62.9 \\
          & ACL~\cite{zhang2022adaptive}   & ResNet50-IBN & 49.4  & \textbf{50.1} & 21.7  & 47.3  & \textbf{76.8} & \textbf{90.6} & 49.3  & 62.7 \\
       %   & IL~\cite{tan2023style}   & ResNet50 & 41.0  & 41.8 & 23.8  & 51.2  & 72.0 & 88.5 & 45.6  & 60.5 \\
          & IL$^{\dag}$~\cite{tan2023style}   & ResNet50-IBN & 40.4  & 41.0 & 23.4  & 49.3  & 71.4 & 87.4 & 45.1  & 59.2 \\
          & GMN(Ours)   & ResNet50-IBN  & \textbf{49.5} & \textbf{50.1} & \textbf{24.8} & \textbf{51.0} & 75.9  & 89.0  & \textbf{50.1} & \textbf{63.4} \\
    \bottomrule
    \end{tabular}%
  \label{tab01}%
  \vspace{-10pt}
\end{table*}% 

% In this part, we firstly introduce the experimental datasets and settings in Section~\ref{sec:EXP-DS}. Then, we compare the proposed method with the state-of-the-art generalizable Re-ID methods in Section~\ref{sec:EXP-CUA}, respectively. To validate the effectiveness of various components in the proposed framework, we conduct ablation studies in Section~\ref{sec:EXP-SS}. Lastly, we further analyze the property of the proposed method in Section~\ref{sec:EXP-FA}.
 In this section, we begin by presenting the experimental datasets and configurations in Section~\ref{sec:EXP-DS}. Following that, we evaluate our proposed method against the current state-of-the-art generalizable Re-ID techniques in Section~\ref{sec:EXP-CUA}. To verify the impact of different modules in our framework, we carry out ablation studies in Section~\ref{sec:EXP-SS}. Finally, we delve deeper into the properties of our approach in Section~\ref{sec:EXP-FA}.
\subsection{Datasets and Experimental Settings}\label{sec:EXP-DS}
 \subsubsection{Datasets} 
% We evaluate our approach on four large-scale image datasets: Market1501~\cite{DBLP:conf/iccv/ZhengSTWWT15}, CUHK-SYSU~\cite{xiao2016end}, MSMT17~\cite{wei2018person} and CUHK03-NP~\cite{DBLP:conf/cvpr/LiZXW14,DBLP:conf/cvpr/ZhongZCL17}. 
%  \textbf{Market1501 (M)} contains 1,501 persons with 32,668 images from six cameras. Among them, $12,936$ images of $751$ identities are used as a training set. For evaluation, there are $3,368$ and $19,732$ images in the query set and the gallery set, respectively. 
%  \textbf{MSMT17 (MS)} is collected from a 15-camera network deployed on campus. The training set contains $32,621$ images of $1,041$ identities. For evaluation, $11,659$ and $82,161$ images are used as query and gallery images, respectively. \textbf{CUHK03-NP (C3)} has an average of 4.8 images per camera for each identity. The dataset provides both manually labeled bounding boxes and DPM-detected bounding boxes. On this dataset, there are $7,365$ training images, and $1,400$ images and $5,332$ images in query set and gallery set are used in the test stage. \textbf{CUHK-SYSU (CS)} has $11,934$ persons with $34,574$ images.
%  Particularly, we divide these four datasets into two parts: three domains as source domains for training and the other one as target domain for test. We adopt the recommended setting in ~\cite{xu2022mimic,zhang2022adaptive}.
%    For all datasets, we employ CMC (\ie, Rank1 (R1), Rank5 (R5) and Rank10 (R10)) accuracy and mAP (\ie, mean Average Precision) for Re-ID evaluation~\cite{DBLP:conf/iccv/ZhengSTWWT15}.

In this part, we provide details about the experimental datasets and configurations. We evaluate our proposed method based on four large datasets: Market1501~\cite{DBLP:conf/iccv/ZhengSTWWT15},  MSMT17~\cite{wei2018person}, CUHK03-NP~\cite{DBLP:conf/cvpr/LiZXW14,DBLP:conf/cvpr/ZhongZCL17} and CUHK-SYSU~\cite{xiao2016end}. The \textbf{Market1501 (M)} dataset consists of $1,501$ persons and $32,668$ images captured from six cameras, and the training set has $12,936$ images of $751$ identities. The \textbf{MSMT17 (MS)} dataset has $32,621$ images of $1,041$ identities in the training set, while $11,659$ and $82,161$ samples are used as query and gallery images, respectively. The \textbf{CUHK03-NP (C3)} dataset has $7,365$ training images, and we use $1,400$ images and $5,332$ images in query set and gallery set for testing. Finally, the \textbf{CUHK-SYSU (CS)} dataset has $11,934$ persons and $34,574$ images. Our experimental setup involves partitioning the four datasets into two sets, where three domains are utilized as source domains during training, and one domain serves as the target domain during testing, which is the same as \cite{xu2022mimic,zhang2022adaptive}. In the test stage, we leverage CMC accuracy (Rank1 (R1), Rank5 (R5) and Rank10 (R10)) and mAP (mean Average Precision) for the Re-ID evaluation~\cite{DBLP:conf/iccv/ZhengSTWWT15}.

\subsubsection{Implementation Details} 
% In this experiment, we use the ResNet-50~\cite{DBLP:conf/cvpr/HeZRS16} with IBN module~\cite{DBLP:conf/eccv/PanLST18}, which is pre-trained on ImageNet~\cite{DBLP:conf/cvpr/DengDSLL009} to initialize the network parameters. For the cross-entropy loss, we employ the label smoothing scheme during the training course. In a batch, the number of IDs and the number of images per person are set as $16$ and $4$ to produce triplets for each domain, respectively.
%  The initial learning rate is $3.5\times 10^{-4}$ and divided by $10$ at the $40$-th and $90$-th epochs, respectively. The proposed model is trained with the Adam optimizer in a total of $120$ epochs. We activate the DP module at 60-th epoch, and the DP module randomly drops 50\% of the feature map on the channel per operation. The size of the input image is $256 \times 128$. For data augmentation, we perform random cropping,
% random flipping and auto-augmentation~\cite{cubuk2018autoaugment}. Besides,  $\lambda$ in Eq.~\ref{eq09} is set as $1.0$. %In all experiments, the 
%  Particularly, we utilize the same setting for all experiments on all datasets in this paper. 

In this study, we initialize the network parameters using ResNet-50~\cite{DBLP:conf/cvpr/HeZRS16} with IBN module~\cite{DBLP:conf/eccv/PanLST18} that is pre-trained on ImageNet~\cite{DBLP:conf/cvpr/DengDSLL009}. In a batch, each domain involves $16$ identities, and we sample $4$ images per identity to produce triplets. The initial learning rate is set to $3.5\times 10^{-4}$, which is divided by $10$ at the $40$-th and $90$-th epochs. The proposed model is trained with the Adam optimizer for a total of $120$ epochs. We activate the DP module at the $60$-th epoch, which randomly drops 50\% of the feature map on the channel per operation. The size of the input image is $256 \times 128$, and we use random cropping, flipping, and auto-augmentation~\cite{cubuk2018autoaugment} for data augmentation. Additionally, we set $\lambda$ in Eq.~\ref{eq09} as $1.0$. We employ the same settings for all experiments on all datasets in this paper.

\subsection{Comparison with State-of-the-art Methods}\label{sec:EXP-CUA}

We compare our method with state-of-the-art (SOTA) methods, including SNR~\cite{DBLP:conf/cvpr/JinLZ0Z20}, QAConv$_{50}$~\cite{DBLP:conf/eccv/LiaoS20}, M$^3$L~\cite{DBLP:conf/cvpr/ZhaoZYLLLS21}, MetaBIN~\cite{DBLP:conf/cvpr/ChoiKJPK21}, META~\cite{xu2022mimic}, IL~\cite{tan2023style} and ACL~\cite{zhang2022adaptive}. The SNR technique utilizes instance normalization to extract identity-relevant features from discarded information and reintegrates them into the network to maintain high discriminability. QAConv generates query-adaptive convolutional kernels dynamically to enable local matching. Both M$^3$L and MetaBIN are meta-learning based methods. META employs domain-specific networks to supplement domain-specific information for each domain. ACL utilizes a shared feature space to extract both domain-invariant and domain-specific features, while also dynamically discovering relationships across different domains. The backbone of all these methods is ResNet-50 combined with instance normalization, and M$^3$L and ACL also employ the IBN structure~\cite{DBLP:conf/eccv/PanLST18}, thus it is absolutely fair in this comparison.

We report experimental results in Tab.~\ref{tab01}. In this table, ``Training sets'' means that we merely use training set of all source domains to train the model, and ``Full Images'' is that we use the training and test set of all source domains to train the model. Note that MetaBIN~\cite{DBLP:conf/cvpr/ChoiKJPK21} and META~\cite{xu2022mimic} use ResNet50-BIN and ResNet50-DSBN as the  backbone, respectively. Particularly, the backbone of these methods include BN and IN components. As seen, our method can obtain the best result when compared with other methods in "M+MS+CS$\rightarrow$C3" and "M+CS+C3$\rightarrow$MS". Besides, although our method is slightly poorer than ACL, the averaged result of our method is best in all tasks, which validates our method is beneficial for improving the generalization capability in unseen domains. For example, our method can outperform ACL by $+1.3\%$ (46.6 vs. 45.3) on mAP in the case based on the training set. It is worth noting that ACL acquires four GPUs to train the model, while our method only needs one GPU. Moreover, the training time of ACL is three times that of our method.

 We also conduct relevant experiments on DukeMTMC-ReID and compared them with some methods mentioned in our paper. The specific results are presented in Tab.~\ref{tab21}. 
Generally speaking, our method achieves better results compared to other methods. It is worth noting that META~\cite{xu2022mimic} uses ResNet50 with DSBN~\cite{chang2019domain} (\ie, using independent normalization layer for each domain) as the backbone due to the characteristic of multiple experts. Besides, IL~\cite{tan2023style} only provides the code based on ResNet-50, thus we replace ResNet-50 with the ResNet50-IBN to run the experiment so as to achieve the fair comparison.

\begin{table*}[htbp]
  \centering
  \caption{Experiments conducted on DukeMTMC-ReID (D). Market1501 (M), MSMT17 (MS) and CUHK03 (C3) are utilized as source domains during training, and DukeMTMC-ReID (D) serves as the target domain during testing. The results use the training set images on each dataset, respectively. Note that the \textbf{bold} is the best result and {\dag} indicates evaluation results based on the code released by the authors.}
    \begin{tabular}{c|c|c|cc|cc|cc|cc}
    \toprule
    \multirow{2}[1]{*}{Setting} & \multirow{2}[1]{*}{Method} & \multirow{2}[1]{*}{~~Backbone~~} & \multicolumn{2}{c|}{~~~M+MS+CS$\rightarrow$D~~~} & \multicolumn{2}{c|}{~~~M+CS+C3$\rightarrow$D~~~} & \multicolumn{2}{c|}{~~~MS+CS+C3$\rightarrow$D~~~} & \multicolumn{2}{c}{Average} \\
\cmidrule{4-11}          &       &       & mAP   & R1  & mAP   & R1 & mAP   & R1  & ~~mAP~~   & ~~R1~~  \\
    \midrule
    \multirow{4}[2]{*}{Training Sets}
          & ~~META~\cite{xu2022mimic}~~  & ~~ResNet50-DSBN~~ & 49.4 & 68.8  & 42.6  & 62.7  & 49.2  & 69.0  & 47.1  & 66.8 \\
          & ACL~\cite{zhang2022adaptive}   &  ResNet50-IBN & \textbf{56.4} & 73.0  & 50.9 & 70.3  & 56.5 & 74.3 & 54.6  & 72.5 \\
          & IL$^{\dag}$ ~\cite{tan2023style}  & ResNet50-IBN & 54.8 & \textbf{73.7}  & 48.3  & 68.5  & 55.7  & 72.8  & 52.9  & 71.7 \\
          & GMN(Ours)   & ResNet50-IBN  & 55.4 & 72.3 & \textbf{52.1} & \textbf{70.6} & \textbf{58.6}  & \textbf{74.8}  & \textbf{55.4} & \textbf{72.6} \\
    % \midrule
    % \multirow{7}[2]{*}{Full Images} & SNR~\cite{DBLP:conf/cvpr/JinLZ0Z20}  & CVPR2020 & 17.5 & 17.1  & 7.7   & 22.0  & 52.4  & 77.8  & 25.9  & 39.0 \\
    %       & QAConv$_{50}$~\cite{DBLP:conf/eccv/LiaoS20} & ECCV2020 & 32.9 & 33.3  & 17.6  & 46.6  & 66.5  & 85.0  & 39.0  & 55.0 \\
    %       & M$^3$L~\cite{DBLP:conf/cvpr/ZhaoZYLLLS21} & CVPR2021 & 35.7 & 36.5  & 17.4  & 38.6  & 62.4  & 82.7  & 38.5  & 52.6 \\
    %       & MetaBIN~\cite{DBLP:conf/cvpr/ChoiKJPK21} & CVPR2021 & 43.0 & 43.1  & 18.8  & 41.2  & 67.2  & 84.5  & 43.0  & 56.3 \\
    %       & META~\cite{xu2022mimic}  & ECCV2022 & 47.1 & 46.2  & 24.4  & 52.1  & 76.5  & 90.5  & 49.3  & 62.9 \\
    %       & ACL~\cite{zhang2022adaptive}   & ECCV2022 & 49.4  & \textbf{50.1} & 21.7  & 47.3  & \textbf{76.8} & \textbf{90.6} & 49.3  & 62.7 \\
    %       & GMN   & This paper  & \textbf{49.5} & \textbf{50.1} & \textbf{24.8} & \textbf{51.0} & 75.9  & 89.0  & \textbf{50.1} & \textbf{63.4} \\
    \bottomrule
    \end{tabular}%
      \label{tab21}%
  \vspace{-10pt}
\end{table*}% 

% To further confirm the superiority of the proposed method, we also perform the experiment on small datasets including PRID~\cite{DBLP:conf/scia/HirzerBRB11}, GRID~\cite{loy2010time} and VIPeR~\cite{DBLP:conf/eccv/GrayT08}. Experimental results are shown in Tab.~\ref{tab11}. As seen in this table, based on the averaged result of three datasets, our method increase mAP and R1 by $1.2\%$ (72.6 vs. 71.4) and $+1.8\%$ (63.1 vs. 61.5) when compared with the currently SOTA ACL. This again validate the value of our method.

To further confirm the superiority of the proposed method, we also performed experiments on small datasets including PRID~\cite{DBLP:conf/scia/HirzerBRB11}, GRID~\cite{loy2010time}, and VIPeR~\cite{DBLP:conf/eccv/GrayT08}. The experimental results are shown in Tab.~\ref{tab11}. As can be seen in this table, based on the averaged results of the three datasets, our method increases mAP and R1 by $1.2\%$ (72.6 vs. 71.4) and $+1.8\%$ (63.1 vs. 61.5) compared to the currently SOTA method ACL. This again validates the value of our method.
\begin{table}[htbp]
  \centering
  \caption{Comparison with SOTA methods on small datasets. All methods are trained on Market1501 (M), MSMT17 (MS), CUHK03 (C3) and CUHK-SYSU (CS).}
    \begin{tabular}{L{1.6cm}|C{0.4cm}C{0.4cm}|C{0.4cm}C{0.4cm}|C{0.4cm}C{0.4cm}|C{0.4cm}C{0.4cm}}
    \toprule
    \multicolumn{1}{c|}{\multirow{2}[1]{*}{Methods}}  & \multicolumn{2}{c|}{PRID} & \multicolumn{2}{c|}{GRID} & \multicolumn{2}{c|}{VIPeR} & \multicolumn{2}{c}{Average} \\
\cmidrule{2-9}          & mAP   & R1    & mAP   & R1    & mAP   & R1    & mAP   & R1 \\
    \midrule
    QAConv$_{50}$~\cite{DBLP:conf/eccv/LiaoS20} & 62.2 & 52.3  & 57.4  & 48.6  & 66.3  & 57.0  & 61.9  & 52.6 \\
    M$^3$L~\cite{DBLP:conf/cvpr/ZhaoZYLLLS21}  & 64.3 & 53.1  & 55.0  & 44.4  & 66.2  & 57.5  & 60.6  & 51.7 \\
    MetaBIN~\cite{DBLP:conf/cvpr/ChoiKJPK21} & 70.8  & 61.2  & 57.9  & 50.2  & 64.3  & 55.9  & 61.1  & 55.8 \\
    META~\cite{xu2022mimic}  & 71.7  & 61.9  & 60.1  & 52.4  & 68.4  & 61.5  & 66.7  & 58.6 \\
    ACL~\cite{zhang2022adaptive}    & 73.4  & 63.0  & \textbf{65.7} & \textbf{55.2} & 75.1  & 66.4  & 71.4  & 61.5 \\
    \midrule
    GMN (Ours)  & \textbf{75.4} & \textbf{66.0} & 64.8  & 54.4  & \textbf{77.7} & \textbf{69.0} & \textbf{72.6} & \textbf{63.1} \\
    \bottomrule
    \end{tabular}%
  \label{tab11}%
   \vspace{-10pt}
\end{table}%

\subsection{Ablation Study}~\label{sec:EXP-SS}
% In this part, we conduct the ablation experiment to validate the effectiveness of each module in our method, as listed in Tab.~\ref{tab02}. ``A'', ``B'' and ``C'' in this table represent metric network, dropout-based perturbation and pair-identity center loss, respectively. As observed in this table, all modules are valuable in all tasks, which can be beneficial to improve the model's generalization. For example, in ``M+MS+CS$\rightarrow$C3'', using the metric network increases $+4.7\%$ (39.5 vs. 34.8) on mAP when compared with baseline. Besides, adding the dropout-based perturbation can further improve mAP by $+1.5\%$ (41.0 vs. 39.5) over ``Baseline+A''. Moreover, ``Baseline+A+B+C'' can obtain the best result in all tasks, which confirms the efficacy of pair-identity center loss.

In this section, we conduct an ablation experiment to evaluate the effectiveness of each module in our proposed method. The results are presented in Tab.~\ref{tab02}, where ``A'', ``B'', and ``C'' represent the metric network, dropout-based perturbation, and pair-identity center loss, respectively. As shown in the table, all modules are valuable and can improve the model's generalization in all tasks. For instance, using the metric network in ``M+MS+CS$\rightarrow$C3'' increases the mAP by $+4.7\%$ (39.5 vs. 34.8) compared to the baseline. Additionally, adding the dropout-based perturbation further improves the mAP by $+1.5\%$ (41.0 vs. 39.5) over ``Baseline+A''. Finally, ``Baseline+A+B+C'' achieves the best results in all tasks, confirming the effectiveness of the pair-identity center loss.
\begin{table*}[htbp]
  \centering
  \caption{Ablation results on Market1501 (M), MSMT17 (MS) and CUHK03 (C3). ``A'' denotes the metric network, ``B'' is the dropout-based perturbation, and ``C'' is the pair-identity center loss. Note that the model is trained on the training set of all source domains.}
    \begin{tabular}{l|cccc|cccc|cccc|cccc}
    \toprule
    \multicolumn{1}{c|}{\multirow{2}[1]{*}{Methods}} & \multicolumn{4}{c|}{M+MS+CS$\rightarrow$C3} & \multicolumn{4}{c|}{M+CS+C3$\rightarrow$MS} & \multicolumn{4}{c|}{MS+CS+C3$\rightarrow$M} & \multicolumn{4}{c}{Average} \\
\cmidrule{2-17}          & mAP   & R1    & R5    & R10   & mAP   & R1    & R5    & R10   & mAP   & R1    & R5    & R10   & mAP   & R1    & R5    & R10 \\
    \midrule
    Baseline & 34.8  & 34.5  & 52.9  & 62.8  & 20.8  & 44.6  & 57.8  & 63.8  & 62.6  & 82.5  & 91.4  & 94.1  & 39.4  & 53.9  & 67.4  & 73.6 \\
    Baseline+A & 39.5  & 40.0  & 59.9  & 71.4  & 22.6  & 48.1  & 61.9  & 67.5  & 68.9  & 85.0  & 93.7  & 96.1  & 43.7  & 57.7  & 71.8  & 78.3 \\
    Baseline+A+B & 41.0  & 40.7  & 63.7  & 74.6  & 23.8  & 50.0  & 64.0  & 69.8  & 71.1  & 86.2  & 94.3  & 96.4  & 45.3  & 59.0  & 74.0  & 80.3 \\
    Baseline+A+B+C & \textbf{43.2} & \textbf{42.1} & \textbf{65.4} & \textbf{75.0} & \textbf{24.4} & \textbf{50.9} & \textbf{64.4} & \textbf{70.3} & \textbf{72.3} & \textbf{87.1} & \textbf{94.5} & \textbf{96.7} & \textbf{46.6} & \textbf{60.0} & \textbf{74.8} & \textbf{80.7} \\
    \bottomrule
    \end{tabular}%
  \label{tab02}%
    %\vspace{-10pt}
\end{table*}%

% Since the pair-identity center loss ($L_{pic}$) consists of $L_{pic+}$ and $L_{pic-}$, we also perform the experiment to validate the effectiveness of two items, as shown in Tab.~\ref{tab03}. As seen in table, we find that both $L_{pic+}$ and $L_{pic-}$ have a positive impact on improving the model's generalization. Based on the averaged result of three tasks, their fused result is better than the independent result on the whole.

To validate the effectiveness of the two components of the pair-identity center loss ($L_{pic}$), namely $L_{pic+}$ and $L_{pic-}$, we conduct an experiment, and the results are shown in Tab.~\ref{tab03}. From the table, it is evident that both $L_{pic+}$ and $L_{pic-}$ have a positive impact on improving the model's generalization. It can be seen that the boost of $L_{pic-}$ is a little more than the boost of $L_{pic+}$. This is due to the fact that the relationship between negative sample pairs is more valuable in the process of metric. Moreover, the fused result of the two components is better than the independent result on the whole, based on the averaged result of three tasks.
\begin{table*}[htbp]
  \centering
  \caption{Ablation experiments of the pair-identity center loss on Market1501 (M), MSMT17 (MS) and CUHK03 (C3).}
    \begin{tabular}{l|cccc|cccc|cccc|cccc}
    \toprule
    \multicolumn{1}{c|}{\multirow{2}[1]{*}{Methods}}   & \multicolumn{4}{c|}{M+MS+CS$\rightarrow$C3} & \multicolumn{4}{c|}{M+CS+C3$\rightarrow$MS} & \multicolumn{4}{c|}{MS+CS+C3$\rightarrow$M} & \multicolumn{4}{c}{Average} \\
\cmidrule{2-17}          & ~~mAP   & R1    & R5    & R10~~   & ~~mAP   & R1    & R5    & R10~~   & ~~mAP   & R1    & R5    & R10~~   & mAP   & R1    & R5    & R10 \\
    \midrule
    None  & 41.0  & 40.7  & 63.7  & 74.6  & 23.8  & 50.0  & 64.0  & 69.8  & 71.1  & 86.2  & 94.3  & 96.4  & 45.3  & 59.0  & 74.0  & 80.3 \\
    $L_{pic+}$   & 41.9  & 42.1  & 63.2  & 74.7  & 24.1  & 50.6  & 64.1  & 69.9  & 71.9  & 86.6  & 94.2  & 96.6  & 46.0  & 59.8  & 73.8  & 80.4 \\
     $L_{pic-}$   & \textbf{43.3} & \textbf{43.1} & 65.0  & 74.0  & 24.3  & 50.7  & \textbf{64.6} & \textbf{70.4} & 72.0  & 86.8  & 94.3  & 96.4  & 46.5  & \textbf{60.2}  & 74.6  & 80.3 \\
     \rowcolor[rgb]{ .749,  .749,  .749}  $L_{pic}$ & 43.2  & 42.1  & \textbf{65.4} & \textbf{75.0} & \textbf{24.4} & \textbf{50.9} & 64.4  & 70.3  & \textbf{72.3} & \textbf{87.1} & \textbf{94.5} & \textbf{96.7} & \textbf{46.6} & 60.0  & \textbf{74.8} & \textbf{80.7} \\
    \bottomrule
    \end{tabular}%
  \label{tab03}%
\end{table*}%

\subsection{Further Analysis}~\label{sec:EXP-FA}
% \textbf{Evaluation on different sampling schemes on negative pairs.} We analyze the impact of different sampling scheme in this part. Here, we generate the negative sample-pairs within each domain and  across different domains, respectively. The experimental results are reported in Tab.~\ref{tab04}. In this paper, we randomly generate negative sample-pairs, \ie, two samples of a negative sample-pair could be from the same domain or different domains. As seen in this table, using the random scheme can obtain the better performance when compared with only inter-domain or intra-domain sampling.

\textbf{Evaluation of Different Sampling Schemes for Negative Pairs.} In this section, we analyze the impact of different sampling schemes for negative pairs. Specifically, we generate negative sample pairs within each domain and across different domains, and evaluate the performance of our method on each scheme. The experimental results are reported in Tab.~\ref{tab04}. In this paper, we randomly generate negative sample pairs, which means that two samples of a negative pair could be from the same domain or different domains. As shown in the table, using the random sampling scheme leads to better performance compared to using only inter-domain or intra-domain sampling.

% Table generated by Excel2LaTeX from sheet 'final'
\begin{table*}[htbp]
  \centering
  \caption{Experimental results of different sampling schemes on negative pairs on Market1501 (M), MSMT17 (MS), CUHK03 (C3) and CUHK-SYSU (CS). ``Inter-domain'' denotes producing negative sample-pairs across different domains. ``Intra-domain''  is producing negative sample-pairs within each domain. In this paper, we randomly yield the negative sample-pairs in a batch.}
    \begin{tabular}{l|cccc|cccc|cccc|cccc}
    \toprule
    \multicolumn{1}{c|}{\multirow{2}[1]{*}{Methods}} & \multicolumn{4}{c|}{M+MS+CS$\rightarrow$C3} & \multicolumn{4}{c|}{M+CS+C3$\rightarrow$MS} & \multicolumn{4}{c|}{MS+CS+C3$\rightarrow$M} & \multicolumn{4}{c}{Average} \\
\cmidrule{2-17}          & mAP   & R1    & R5    & R10   & mAP   & R1    & R5    & R10   & mAP   & R1    & R5    & R10   & mAP   & R1    & R5    & R10 \\
    \midrule
    Inter-domain~~ & 42.1  & 41.8  & 64.0  & 74.7  & 24.3  & \textbf{50.9} & 64.3  & 70.2  & \textbf{72.3} & 86.9  & \textbf{94.5} & 96.6  & 46.2  & 59.9  & 74.3  & 80.5 \\
    Intra-domain & 42.5  & 41.8  & 64.9  & \textbf{75.1} & 24.1  & 50.2  & 63.7  & 69.4  & 71.5  & 86.6  & \textbf{94.5} & 96.5  & 46.0  & 59.5  & 74.4  & 80.3 \\
    \rowcolor[rgb]{ .749,  .749,  .749} Random & \textbf{43.2} & \textbf{42.1} & \textbf{65.4} & 75.0  & \textbf{24.4} & \textbf{50.9} & \textbf{64.4} & \textbf{70.3} & \textbf{72.3} & \textbf{87.1} & \textbf{94.5} & \textbf{96.7} & \textbf{46.6} & \textbf{60.0} & \textbf{74.8} & \textbf{80.7} \\
    \bottomrule
    \end{tabular}%
  \label{tab04}%
  \vspace{-10pt}
\end{table*}%

  \textbf{Comparison between center (C) loss and pair-identity center (PIC) loss.} We compare our PIC loss with the center loss~\cite{DBLP:conf/eccv/WenZL016} that aims to make all sample's features with the same identity become consistent. The experimental results are given in Tab.~\ref{tab12}. According to the comparison, we find that the proposed PIC loss can obtain better performance than the typical center loss, which thanks to the fact that our task is a metric task and the PIC loss can make the sample-pair features with the same identity consistent.
% Table generated by Excel2LaTeX from sheet 'final'
\begin{table}[htbp]
  \centering
  \caption{Experimental results of center (C) loss and pair-identity center (PIC) loss.}
    \begin{tabular}{L{1.3cm}|C{0.45cm}C{0.45cm}C{0.45cm}C{0.45cm}|C{0.45cm}C{0.45cm}C{0.45cm}C{0.45cm}}
    \toprule
    \multicolumn{1}{c|}{\multirow{2}[1]{*}{Methods}}   & \multicolumn{4}{c|}{M+MS+CS$\rightarrow$C3} & \multicolumn{4}{c}{MS+CS+C3$\rightarrow$M} \\
\cmidrule{2-9}          & mAP   & R1    & R5    & R10   & mAP   & R1    & R5    & R10 \\
    \midrule
    C loss~\cite{DBLP:conf/eccv/WenZL016} & 41.3  & 40.7  & 63.8  & 73.9  & 71.4  & 87.0  & 94.3  & 96.5 \\
    PIC loss & \textbf{43.2} & \textbf{42.1} & \textbf{65.4} & \textbf{75.0} & \textbf{72.3} & \textbf{87.1} & \textbf{94.5} & \textbf{96.7} \\
    \bottomrule
    \end{tabular}%
  \label{tab12}%
  \vspace{-10pt}
\end{table}%

\begin{figure}[t]
\centering
\includegraphics[width=8.5cm]{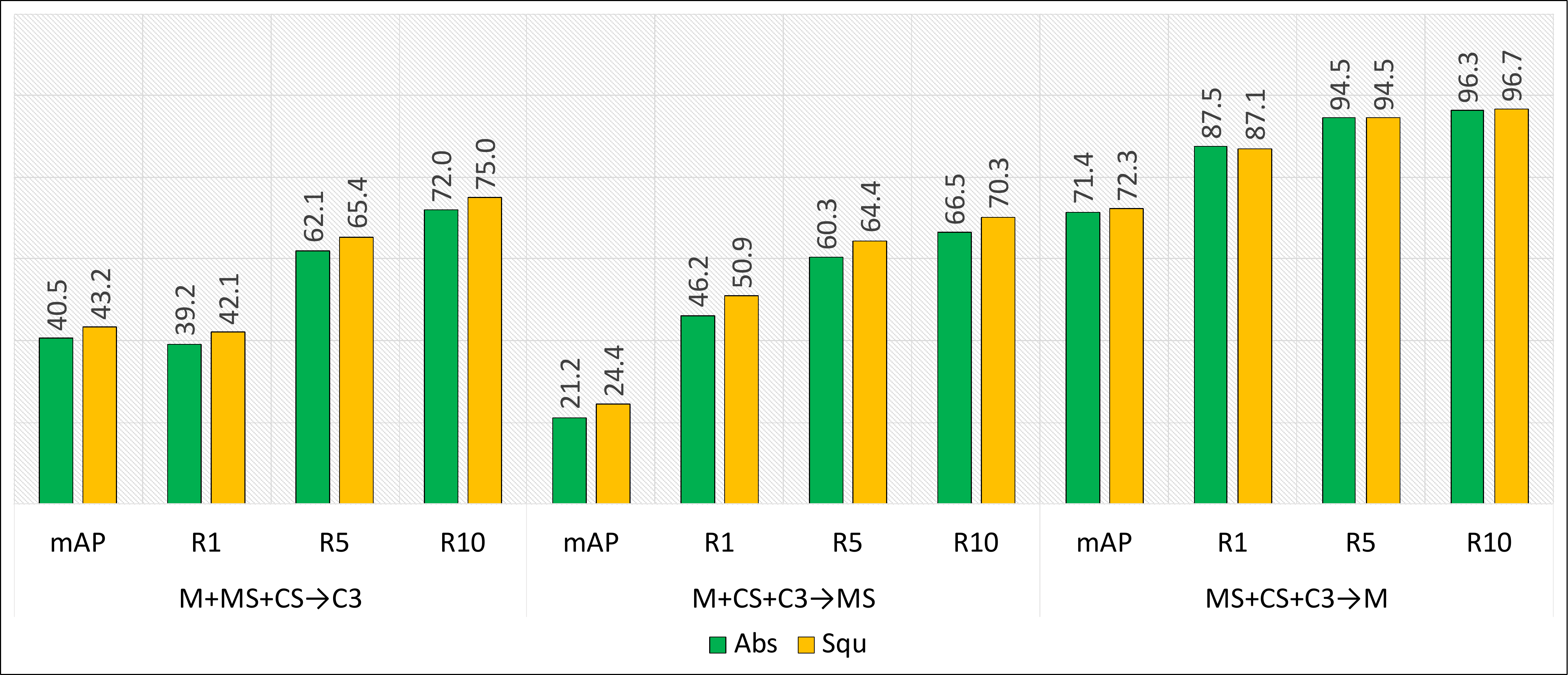}
\caption{Comparison of different schemes on generating sample-pair features in three tasks. In this figure, ``Squ'' is the operation in Eq.~\ref{eq01}. ``Abs'' is the absolute-value operation, \ie, $|a-b|$.}
\label{fig01}
\vspace{-15pt}% µ÷Õû¼ä¾à
\end{figure}

\textbf{Evaluation on different schemes on generating sample-pair features.} We conduct an analysis of the impact of different schemes on generating sample-pair features, as presented in Fig.~\ref{fig01}. Our findings indicate that the dot-product operation in Eq.~\ref{eq01} outperforms the absolute-value operation. Moreover, we also explore various strategies, including absolute values (ABS), multiplication (MUL) and addition (ADD). We follow Protocol 2 as outlined in the paper for both training and testing. The specific experimental results are presented in Tab.~\ref{tab24}.
\begin{table*}[htbp]
  \centering
  \caption{Experimental results for different strategies in creating the sample-pair space on Market1501 (M), MSMT17 (MS) and CUHK03 (C3).}
    \begin{tabular}{c|cccc|cccc|cccc|cccc}
    \toprule
    \multirow{2}[1]{*}{Method} & \multicolumn{4}{c|}{M+MS+CS$\rightarrow$C3} & \multicolumn{4}{c|}{M+CS+C3$\rightarrow$MS} & \multicolumn{4}{c|}{MS+CS+C3$\rightarrow$M} & \multicolumn{4}{c}{Average} \\
\cmidrule{2-17}          & ~~mAP~~   & R1    & R5    & ~~R10~~   & ~~mAP~~   & R1    & R5    & ~~R10~~   & ~mAP~   & R1    & R5    & ~R10~   & mAP   & R1    & R5    & R10 \\
    \midrule
    ABS  & 40.5  & 39.2  & 62.1  & 72.0  & 21.2  & 46.2  & 60.3  & 66.5  & 71.4  & \textbf{87.5}  & 94.5  & 96.7  & 44.4  & 57.6  & 72.3  &  78.3 \\
    MUL & 41.0  & 41.3  & 61.7  & 71.9  & 21.7  & 45.9  & 60.6  & 66.4  & 69.9  & 85.4  & 93.8  & 96.1  & 44.2  & 57.5  & 72.0  & 78.1 \\
    ADD & 7.0  & 5.2  & 13.4  & 20.3  & 8.9  & 18.2  & 33.1  & 40.8  & 22.3  & 27.8  & 52.0  & 64.1  & 12.7  & 17.1  & 32.8  & 41.7 \\
    \rowcolor[rgb]{ .749,  .749,  .749} Ours & \textbf{43.2} & \textbf{42.1} & \textbf{65.4} & \textbf{75.0} & \textbf{24.4} & \textbf{50.9} & \textbf{64.4} & \textbf{70.3} & \textbf{72.3} & 87.1 & \textbf{94.5} & \textbf{96.7} & \textbf{46.6} & \textbf{60.0} & \textbf{74.8} & \textbf{80.7} \\
    \bottomrule
    \end{tabular}%
  \label{tab24}%
    %\vspace{-10pt}
\end{table*}%

\textbf{Analysis of hyper-parameter sensitive.} In Eq.~\ref{eq09}, there is a hyper-parameter $\lambda$ to balance the pair-identity center loss and other losses. We conducted an experiment to analyze the effect of this hyper-parameter, as shown in Tab.~\ref{tab05}. As observed in the table, when the hyper-parameter is set to a large or small value, the performance slightly drops compared to when $\lambda=1$. Overall, the best performance is achieved when $\lambda$ is set to 1. Therefore, in this paper, we set $\lambda$ as 1 in all experiments.

% Table generated by Excel2LaTeX from sheet 'final'
\begin{table*}[htbp]
  \centering
  \caption{Experimental results of different $\lambda$ in Eq.~\ref{eq09} on Market1501 (M), MSMT17 (MS) and CUHK03 (C3).}
    \begin{tabular}{c|cccc|cccc|cccc|cccc}
    \toprule
    \multirow{2}[1]{*}{$\lambda$} & \multicolumn{4}{c|}{~~~M+MS+CS$\rightarrow$C3~~~} & \multicolumn{4}{c|}{~~~M+CS+C3$\rightarrow$MS~~~} & \multicolumn{4}{c|}{~~~MS+CS+C3$\rightarrow$M~~~} & \multicolumn{4}{c}{Average} \\
\cmidrule{2-17}          & ~~mAP   & R1    & R5    & R10~~   & ~~mAP   & R1    & R5    & R10~~   & ~~mAP   & R1    & R5    & R10~~   & ~~mAP   & R1    & R5    & R10~~ \\
    \midrule
    ~~0.1~~   & 41.5  & 41.2  & 64.5  & 74.8  & 24.0  & 50.5  & 64.2  & 70.0  & 71.5  & 86.4  & 94.3  & 96.5  & 45.7  & 59.4  & 74.3  & 80.4 \\
    0.5   & 41.9  & 42.1  & 63.9  & 74.3  & \textbf{24.4} & \textbf{50.9} & \textbf{64.6} & 70.2  & 72.1  & \textbf{87.4} & \textbf{94.5} & 96.5  & 46.1  & 60.1  & 74.3  & 80.3 \\
    \rowcolor[rgb]{ .749,  .749,  .749} 1.0   & 43.2  & 42.1  & \textbf{65.4} & 75.0  & \textbf{24.4} & \textbf{50.9} & 64.4  & \textbf{70.3} & 72.3  & 87.1  & \textbf{94.5} & \textbf{96.7} & \textbf{46.6} & 60.0  & \textbf{74.8} & \textbf{80.7} \\
    1.5   & \textbf{43.3} & \textbf{42.6} & 65.2  & \textbf{75.1} & 24.2  & \textbf{50.9} & 64.2  & 70.0  & \textbf{72.4} & 87.2  & \textbf{94.5} & 96.6  & \textbf{46.6} & \textbf{60.2} & 74.6  & 80.6 \\
    2.0   & 43.2  & 42.7  & 64.6  & 74.6  & 24.0  & 50.6  & 64.0  & 69.8  & \textbf{72.4} & 87.3  & \textbf{94.5} & \textbf{96.7} & 46.5  & \textbf{60.2} & 74.4  & 80.4 \\
    \bottomrule
    \end{tabular}%
  \label{tab05}%
  \vspace{-15pt}
\end{table*}%

\textbf{Comparison between feature based and M-Net based evaluations.} In the test stage, we compared the feature-based evaluation (features from backbone) with the Metric-Net based evaluation using the model trained by our method, and the experimental results are presented in Tab.~\ref{tab06}. As can be seen in the table, our method achieved a significant improvement over the baseline (reported in Tab.~\ref{tab02}) under the same evaluation scheme (\ie, the feature-based method). For example, in the ``MS+CS+C3$\rightarrow$M'' setting, our method achieved an mAP of 70.3, which is $+7.7\%$ better than the baseline (62.6), indicating that our method enhances the feature's discrimination and generalization. Moreover, the use of Metric-Net for evaluation in the test stage further improved the performance.
%Based on the model trained by our method, we compare the feature based evaluation with the M-Net based evaluation in the test stage. We report experimental results in Tab.~\ref{tab06}. As seen in this table, compared with the baseline in Tab.~\ref{tab02}, our method can obtain obvious improvement under the same evaluation scheme (\ie, the feature based method). For example, in ``MS+CS+C3$\rightarrow$M'', our method can significantly improve the baseline by $+7.7\%$ (70.3 vs. 62.6) on mAP, which shows that using our method can also enhance the feature's discrimination and generalization. Moreover, if we use the Metric-Net for evaluation in the test stage, we can obtain the better result.
% Table generated by Excel2LaTeX from sheet 'final'
\begin{table}[htbp]
  \centering
  \caption{Experimental results of different evaluation methods in the test stage. ``Feature'' denotes using features extracted by our model to evaluate performance. ``M-Net'' is our method uses the metric net to evaluate the performance.}
    \begin{tabular}{l|cccc|cccc}
    \toprule
    \multicolumn{1}{c|}{\multirow{2}[1]{*}{Methods}}  & \multicolumn{4}{c|}{M+MS+CS$\rightarrow$C3} & \multicolumn{4}{c}{MS+CS+C3$\rightarrow$M} \\
\cmidrule{2-9}          & mAP   & R1    & R5    & R10   & mAP   & R1    & R5    & R10 \\
    \midrule
    Feature & 42.2  & \textbf{42.4} & 63.1  & 73.0  & 70.3  & 86.4  & 94.0  & 96.2 \\
    M-Net & \textbf{43.2} & 42.1  & \textbf{65.4} & \textbf{75.0} & \textbf{72.3} & \textbf{87.1} & \textbf{94.5} & \textbf{96.7} \\
    \bottomrule
    \end{tabular}%
  \label{tab06}%
  \vspace{-5pt}
\end{table}%

\textbf{Analysis of the test time of different evaluation schemes.} In this section, we evaluate the test time of different evaluation schemes, as shown in Tab.~\ref{tab07}. As observed in the table, on CUHK03 and Market1501 datasets, although the Metric-Net based evaluation requires more time than feature-based evaluation, the increase in test time is relatively small. Moreover, on the larger-scale dataset, MSMT17, the Metric-Net based evaluation is slightly faster compared to the feature-based evaluation. This is because the operation of obtaining similarity through M-Net is slightly slower than the original feature-only test method, but there is no need to calculate the distance between features during the test, so the M-Net based evaluation is faster than feature-based evaluation for testing large-scale datasets. When analyzing the testing time under different evaluation strategies, we consistently utilized GPU (GeForce RTX 3090) for the experiments. Given the consideration of algorithmic complexity, our analysis primarily focuses on two aspects: Space Complexity and Time Complexity. Regarding space complexity, the parameter count of the backbone (ResNet-50) is 25.8M, while the parameter count of our metric network is 1.18M, accounting for only 4.5\% of the backbone's parameter count. The difference is relatively small. Regarding time complexity, our analysis focuses on forward propagation and similarity calculation. First, for forward propagation, the original backbone has a time complexity of O($f_{forward}$)), while the metric network has a time complexity of O($f_{GMN}$). As for similarity calculation, our metric network can directly compute similarity, eliminating the need for calculating a similarity matrix based on Euclidean distances between features. In contrast, the original method requires computing a similarity matrix with a complexity of O($f_{Sim}$). Therefore, the time complexity of the feature-based method is O($f_{forward}$) + O($f_{Sim}$), while our method is O($f_{forward}$) + O($f_{GMN}$). Please refer to the below formulas for a detailed estimation of the time complexity.

\begin{equation*}
  \begin{aligned}
  O(N) \sim O(f_{forward}) \sim O(NlogK),
  \end{aligned}
 % \label{eq01}
\end{equation*}
\begin{equation*}
  \begin{aligned}
  O(N) \sim O(f_{forward}) + O(f_{GMN}) \sim O(Nlog(K+k)),
  \end{aligned}
  %\label{eq02}
\end{equation*}
\begin{equation*}
  \begin{aligned}
   O(f_{Sim}) \sim O(n*n),
  \end{aligned}
  %\label{eq03}
\end{equation*}
\begin{equation*}
  \begin{aligned}
  O(n*n) + O(N) \sim O(f_{forward}) + O(f_{Sim}) \\ \sim O(n*n) + O(NlogK),
  \end{aligned}
  %\label{eq04}
\end{equation*}

where \(K\) is the parameter count of the backbone, \(k\) is the parameter count of the metric network, \(N\) is the product of all dimensions of the input images, equal to the number of images multiplied by height multiplied by width multiplied by the number of channels, and \(n\) is the number of images multiplied by the output dimensions of the backbone. From this, we can observe that the time complexity of the feature-based method and our method is close. 
%In this part, we evaluate the test time of different evaluation schemes, as listed in Tab.~\ref{tab07}. As seen in table, on CUHK03 and Market1501, although the test based on the metric-net needs more time than feature based evaluation, the added test time is large. Besides, on the larger-scale dataset (\ie, MSMT17), the metric-net based evaluation is slightly quick compare with the feature based evaluation.
\begin{table}[htbp]
  \centering
  \caption{The test time of different evaluation schemes. In this table, ``s'' is second, and ``m'' is minute.}
    \begin{tabular}{cc|cc|cc}
    \toprule
    \multicolumn{2}{c|}{CUHK03} & \multicolumn{2}{c|}{Market1501} & \multicolumn{2}{c}{MSMT17} \\
    \midrule
    ~~Feature & M-Net~~   & ~~Feature & M-Net~~   & ~~Feature & M-Net~~ \\
    \midrule
    13s   & 19s   & 22s   & 33s   & 6m22s & 6m7s \\
    \bottomrule
    \end{tabular}%
  \label{tab07}%
   \vspace{-5pt}
\end{table}%

\textbf{Analysis of different schemes used for the dropout-based perturbation.} Here, we conducted an experiment to evaluate the impact of different schemes used for the dropout-based perturbation, as listed in Tab.~\ref{tab08}. In this experiment, we applied the dropout-based perturbation at the beginning of the training stage and at the middle of the process, respectively. According to the results in the table, we observed that using the dropout-based perturbation in the middle of the training stage produced better results. In contrast, if we applied this scheme at the beginning, the perturbation could become noise to the initialized model. If we train the model after some epochs, the perturbation could be beneficial for the trained model to further enhance its generalization.
%Here, we conduct the experiment to evaluate the impact of different schemes used for the dropout-based perturbation, as listed in Tab.~\ref{tab08}. In this experiment, we launch the dropout-based perturbation at the begin of the training stage and at the middle process, respectively. According to the result in this table, we observe that using the dropout-based perturbation at the middle of the training stage can produce the better result. In contrast, if we employ this scheme at the begin, the perturbation could become noise to the initialized model. If we train the model after some epochs, the perturbation could be beneficial for the trained model to further enhance the model's generalization.
\begin{table}[htbp]
  \centering
  \caption{Experimental results of different schemes used for the dropout-based perturbation. ``start'' denotes that we use the dropout-based perturbation scheme at the begin of the training stage. ``middle'' is that we use the scheme in the middle epoch of the training process.}
    \begin{tabular}{l|cccc|cccc}
    \toprule
    \multicolumn{1}{c|}{\multirow{2}[1]{*}{Method}}   & \multicolumn{4}{c|}{M+MS+CS$\rightarrow$C3} & \multicolumn{4}{c}{MS+CS+C3$\rightarrow$M} \\
\cmidrule{2-9}          & mAP   & R1    & R5    & R10   & mAP   & R1    & R5    & R10 \\
    \midrule
    Start & 39.8  & 38.4  & 60.8  & 71.8  & 69.7  & 86.2  & 94.2  & 96.1 \\
    \rowcolor[rgb]{ .749,  .749,  .749} Middle & \textbf{43.2} & \textbf{42.1} & \textbf{65.4} & \textbf{75.0} & \textbf{72.3} & \textbf{87.1} & \textbf{94.5} & \textbf{96.7} \\
    \bottomrule
    \end{tabular}%
  \label{tab08}%
  \vspace{-5pt}
\end{table}%

\textbf{Evaluation on different positions of the dropout-based perturbation.} As is well known, ResNet consists of four blocks. In this experiment, we evaluate the impact of inserting the dropout-based perturbation at different positions within the ResNet architecture. The results are reported in Tab.~\ref{tab09}. According to the results, when the perturbation module is inserted after Block3, the performance is the best. When it appears in the shallow layers, it may have a small impact on the Metric-Net. On the other hand, in the deep layers, it may destroy the semantic information. Therefore, inserting the drop-based perturbation module in the middle position produces satisfactory performance.%As known, the ResNet contains four blocks. We evaluate the impact of different positions of the dropout-based perturbation. The experimental results are reported in Tab.~\ref{tab09}. According to the result, when the module is inserted after Block3, the performance is the best. When it appears in the shallow layers, it may have a small impact on the metric-net. Meanwhile, in the deep layers, it may destroy the semantic information. Therefore, we put it into the middle position can produce the satisfied performance.
\begin{table}[htbp]
  \centering
  \caption{Experimental results of different positions of the dropout-based perturbation. ``Block1'' represents that inserting the module after block1 of ResNet.}
    \begin{tabular}{l|cccc|cccc}
    \toprule
  \multicolumn{1}{c|}{\multirow{2}[1]{*}{Position}}   & \multicolumn{4}{c|}{M+MS+CS$\rightarrow$C3} & \multicolumn{4}{c}{MS+CS+C3$\rightarrow$M} \\
\cmidrule{2-9}          & mAP   & R1    & R5    & R10   & mAP   & R1    & R5    & R10 \\
    \midrule
    Block1 & 40.8  & 41.0  & 61.2  & 72.9  & 70.1  & 85.4  & 94.0  & 95.9 \\
    Block2 & 40.6  & 39.8  & 61.9  & 73.2  & 71.1  & 86.2  & 94.4  & 96.3 \\
 \rowcolor[rgb]{ .749,  .749,  .749}    Block3 & \textbf{43.2} & \textbf{42.1} & \textbf{65.4} & \textbf{75.0} & \textbf{72.3} & \textbf{87.1} & \textbf{94.5} & \textbf{96.7} \\
    Block4 & 39.7  & 39.5  & 60.1  & 71.1  & 69.6  & 85.3  & 93.7  & 96.0 \\
    \bottomrule
    \end{tabular}%
  \label{tab09}%
\end{table}%

\textbf{Performance on source domains.} Here, we also report results on source domains, as shown in Tab.~\ref{tab10}. As seen in this table, our method performs slightly worse than the baseline on the source domains. However, for the domain generalization task, overfitting to source domains should be avoided. Therefore, our method can be considered as an effective approach to alleviate the overfitting to source domains.%Here, we also report results on source domains, as listed in Tab.~\ref{tab10}. As seen in this table, our method is slightly poorer than baseline. For the domain generalization task, we can consider it as overfitting to source domains. From this perspective, our method can alleviate the overfitting to source domains.

\begin{figure}
\centering
\includegraphics[width=8.5cm]{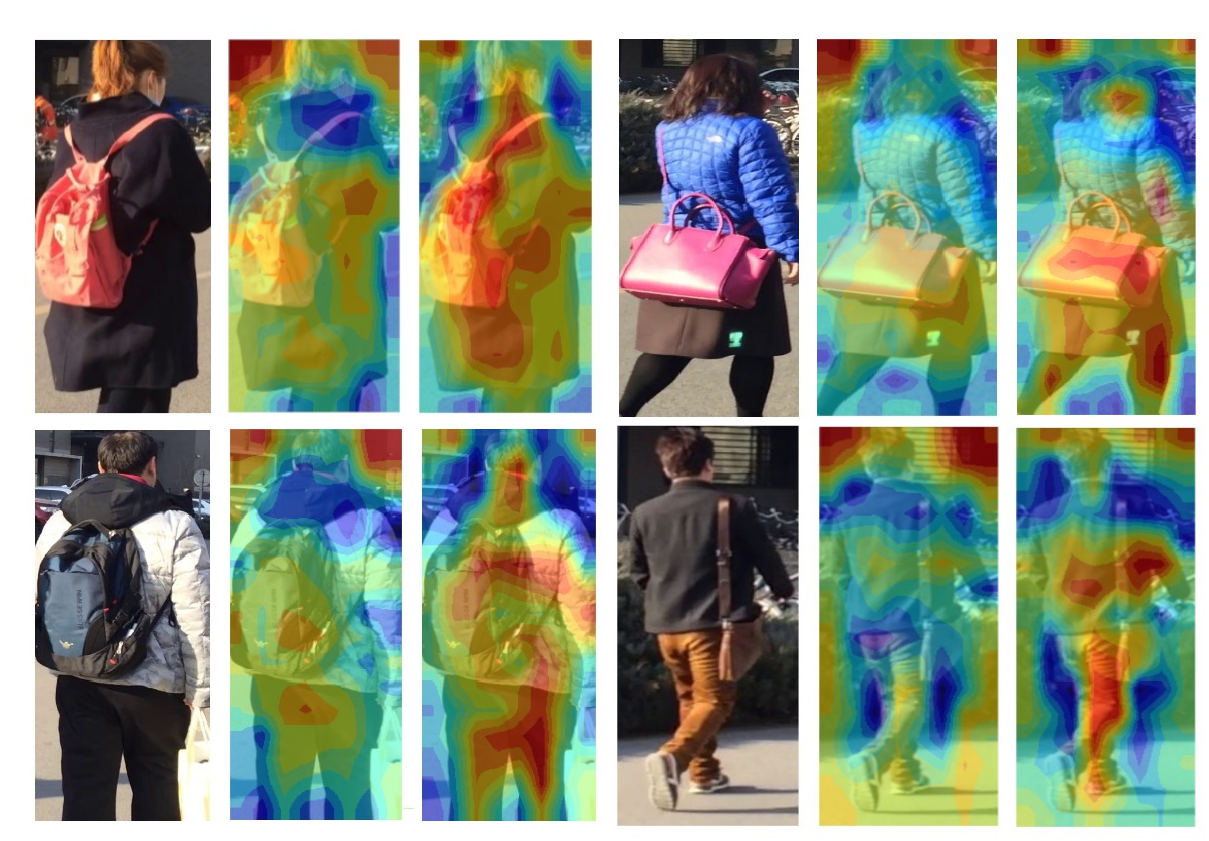}
\caption{Visualization of activation regions in the unseen domain. The first column is the original image, the second column is the activation map from the baseline, and the last column is the activation map from our method.}
\label{fig02}
 \vspace{-15pt}
\end{figure}

\textbf{Visualization of activation regions.} In Fig.~\ref{fig02}, we can see the activation regions produced by the baseline method and our proposed method. The activated regions of three different images for each person in the dataset are shown. As can be observed, the activation regions produced by our method are larger than those produced by the baseline method. This indicates that our method is able to focus on larger regions of the input images, improving the model's generalization ability.%We visualize the activation regions, as illustrated in Fig.~\ref{fig02}. As observed, using our method can produce larger activation region than the baseline method, which means that our method focuses on large regions so as to obtain the good generalization. 

\textbf{Comparison to some big ReID models.} Although big ReID models often excel at learning more intricate and abstract feature representations, crucial for handling the intricate variations in person re-identification. It is noteworthy that, from the perspective of domain generalization, intricate feature representations may not adeptly address differences in unseen domains. Our proposed method enables the network to learn the fundamental distinctions among pedestrians, thereby more effectively mitigating domain gaps. Regarding the big models you mentioned, we utilize the pre-trained models provided by LUPerson~\cite{fu2021unsupervised} and APTM~\cite{yang2023towards} to extract features and evaluate their generalization performance. We conduct the following experiments in Tab.~\ref{tab23}. We observe that APTM offers multiple pre-trained models, and some of these models are trained on datasets that align with the same dataset used for DG-ReID, which does not meet the requirements of DG-ReID. Therefore, we opt for the pre-trained model from APTM on the PA-100K dataset~\cite{liu2017hydraplus} to serve as the testing model. Note that 
the results for LUPerson and APTM are directly obtained by testing the features, whereas the results for our method are derived using the Protocol 2 as outlined in our paper.

% \newcolumntype{L}[1]{>{\PreserveBackslash\raggedright}p{#1}}
\begin{table*}[htbp]
  \centering
  \caption{Experimental results of using pretrained model from LUPerson(LUP)~\cite{fu2021unsupervised}, APTM~\cite{yang2023towards} and our method on Market1501 (M), MSMT17 (MS) and CUHK03 (C3). }
    \begin{tabular}{c|cccc|cccc|cccc|cccc}
    \toprule
    \multirow{2}[1]{*}{Method} & \multicolumn{4}{c|}{M+MS+CS$\rightarrow$C3} & \multicolumn{4}{c|}{M+CS+C3$\rightarrow$MS} & \multicolumn{4}{c|}{MS+CS+C3$\rightarrow$M} & \multicolumn{4}{c}{Average} \\
\cmidrule{2-17}          & ~mAP~   & R1    & R5    & ~R10~   & ~mAP~   & R1    & R5    & ~R10~   & ~mAP~   & R1    & R5    & ~R10~   & ~mAP~   & R1    & R5    & ~R10~ \\
    \midrule
    LUP~\cite{fu2021unsupervised}  & 11.8  & 10.1  & 20.8  & 28.6  & 1.9  & 4.6  & 9.7  & 13.1  & 19.0  & 40.1  & 60.3  & 68.6  & 10.9  & 18.3  & 30.3  & 36.8 \\
    ~~APTM~\cite{yang2023towards}~~ & 1.5  & 1.4  & 2.9  & 3.6  & 3.5  & 16.5  & 26.6  & 31.4  & 11.0  & 33.2  & 50.6  & 58.5  & 5.3  & 17.0  & 20.3  & 31.2 \\
    Ours & \textbf{43.2} & \textbf{42.1} & \textbf{65.4} & \textbf{75.0} & \textbf{24.4} & \textbf{50.9} & \textbf{64.4} & \textbf{70.3} & \textbf{72.3} & \textbf{87.1} & \textbf{94.5} & \textbf{96.7} & \textbf{46.6} & \textbf{60.0} & \textbf{74.8} & \textbf{80.7} \\
    \bottomrule
    \end{tabular}%
  \label{tab23}%
\end{table*}%

\section{Conclusion}\label{s-conclusion}
In this paper, we propose a generalizable metric network (GMN) to solve the domain generalization person Re-ID task, which can effectively alleviate the influence of the domain gap between the source domain and the unseen target domain. The GMN includes a metric network, the dropout-based perturbation, and the pair-identity center loss. For the metric network, we construct the sample-pair features to train it, and use it in the test stage. To further improve the model's generalization, we introduce the dropout-based perturbation to enrich the diversity of sample-pair features. Besides, we also design a pair-identity center loss to obtain the discriminative features.  The GMN's effectiveness is supported by numerous experiments conducted on multiple datasets.

\begin{table}[htbp]
  \centering
  \caption{Experimental results on source domains.}
    \begin{tabular}{c|cccc|cccc}
    \toprule
    \multirow{2}[1]{*}{Test Set} & \multicolumn{4}{c|}{M+MS+CS$\rightarrow$C3} & \multicolumn{4}{c}{MS+CS+C3$\rightarrow$M} \\
\cmidrule{2-9}          & mAP   & R1    & R5    & R10   & mAP   & R1    & R5    & R10 \\
    \midrule
          & \multicolumn{8}{c}{Ours} \\
    \midrule
    MS & \cellcolor[rgb]{ .988,  .894,  .839}57.3 & \cellcolor[rgb]{ .988,  .894,  .839}80.6 & \cellcolor[rgb]{ .988,  .894,  .839}89.5 & \cellcolor[rgb]{ .988,  .894,  .839}92.1 & \cellcolor[rgb]{ .867,  .922,  .969}57.9 & \cellcolor[rgb]{ .867,  .922,  .969}81.0 & \cellcolor[rgb]{ .867,  .922,  .969}90.0 & \cellcolor[rgb]{ .867,  .922,  .969}92.6 \\
    C3 &    -   &    -   &    -   &  -     & \cellcolor[rgb]{ 1,  .902,  .6}69.3 & \cellcolor[rgb]{ 1,  .902,  .6}70.5 & \cellcolor[rgb]{ 1,  .902,  .6}85.9 & \cellcolor[rgb]{ 1,  .902,  .6}92.1 \\
    M & \cellcolor[rgb]{ .776,  .878,  .706}85.8 & \cellcolor[rgb]{ .776,  .878,  .706}93.7 & \cellcolor[rgb]{ .776,  .878,  .706}97.2 & \cellcolor[rgb]{ .776,  .878,  .706}98.3 &   -    &   -    &   -    & - \\
    \midrule
          & \multicolumn{8}{c}{Baseline} \\
    \midrule
    MS & \cellcolor[rgb]{ .988,  .894,  .839}58.8 & \cellcolor[rgb]{ .988,  .894,  .839}82.6 & \cellcolor[rgb]{ .988,  .894,  .839}90.7 & \cellcolor[rgb]{ .988,  .894,  .839}92.9 & \cellcolor[rgb]{ .867,  .922,  .969}58.1 & \cellcolor[rgb]{ .867,  .922,  .969}81.8 & \cellcolor[rgb]{ .867,  .922,  .969}90.4 & \cellcolor[rgb]{ .867,  .922,  .969}92.9 \\
    C3 &   -    &  -     &   -    &    -   & \cellcolor[rgb]{ 1,  .902,  .6}69.4 & \cellcolor[rgb]{ 1,  .902,  .6}71.3 & \cellcolor[rgb]{ 1,  .902,  .6}85.7 & \cellcolor[rgb]{ 1,  .902,  .6}91.4 \\
    M & \cellcolor[rgb]{ .776,  .878,  .706}87.6 & \cellcolor[rgb]{ .776,  .878,  .706}95.1 & \cellcolor[rgb]{ .776,  .878,  .706}98.1 & \cellcolor[rgb]{ .776,  .878,  .706}99.0 &  -     &    -   &   -    &  -\\
    \bottomrule
    \end{tabular}%
  \label{tab10}%
  \vspace{-10pt}
\end{table}%


%\section*{Acknowledgment}

\ifCLASSOPTIONcaptionsoff
  \newpage
\fi

\ifCLASSOPTIONcaptionsoff
  \newpage
\fi
%\normalem
\bibliographystyle{IEEEtran}
\bibliography{sigproc}

 \vfill

% \enlargethispage{-5in}

\end{document}